\documentclass{article} 
\usepackage{titletoc}
\usepackage{float}

\usepackage{microtype}
\usepackage{graphicx}
\usepackage{wrapfig}
\usepackage{adjustbox}

\usepackage{booktabs} 
\usepackage[table]{xcolor} 
\usepackage{enumitem}

\colorlet{goodcolor}{blue!70!black}
\definecolor{dimred}{RGB}{150,40,40}

\usepackage{hyperref}

\usepackage[preprint]{neurips_2024}

\usepackage{comment}
\usepackage{verbatim} 
\usepackage{amssymb,amsmath,amsthm}
\usepackage{mathrsfs}
\usepackage{algorithm}
\usepackage{algorithmic}
\usepackage{bm}
\usepackage{pgfplots}

\usepackage{graphicx,booktabs,multirow}
\usepackage{subcaption}

\usepackage[capitalize,noabbrev]{cleveref}

\usepackage{tabularx}
\usepackage{makecell}
\usepackage{pifont}

\usepackage[most]{tcolorbox}
\tcbuselibrary{breakable}

\usepackage{multirow}
\usepackage{multicol}
\usepackage{colortbl}
\definecolor{my_green}{RGB}{51,102,0}
\definecolor{my_purple}{RGB}{160, 43, 147}
\definecolor{my_blue}{RGB}{15, 158, 213}

\definecolor{promptbg}{gray}{0.97}       
\definecolor{promptborder}{gray}{0.85}
\definecolor{prompttext}{RGB}{50,50,50}  
\definecolor{softred}{RGB}{46,89,132}

\definecolor{table-blue}{RGB}{173, 216, 230}
\definecolor{row-highlight}{RGB}{205, 220, 236}

\definecolor{qwen-header}{RGB}{235, 240, 248}
\definecolor{llama-header}{RGB}{248, 244, 235}
\definecolor{section-gray}{RGB}{245, 245, 245}
\definecolor{group-header}{RGB}{230, 235, 245}

\definecolor{zhz_gray}{rgb}{0.8,0.8,0.8}
\definecolor{darkgreen}{rgb}{0.0, 0.5, 0.0} 
\definecolor{darkred}{rgb}{0.5, 0.0, 0.0}   
\theoremstyle{definition}
\newtheorem{theorem}{Theorem}[section]
\newtheorem{proposition}[theorem]{Proposition}

\theoremstyle{definition}

\newtheorem{remark}[theorem]{Remark}

\usepackage{mathtools}

\title{Iterative Critique-and-Routing Controller for Multi-Agent Systems with Heterogeneous LLMs}

\author{
  Wenzhi Fang$^{1}$,
  Liangqi Yuan$^{1}$,
  Guangchen Lan$^{1}$, \\
  \textbf{Dong-Jun Han}$^{2}$,
 \textbf{Christopher G. Brinton}$^{1}$
}

\begin{document}

\maketitle

\begin{abstract}
Multi-agent large language model (LLM) systems often rely on a controller to coordinate a pool of heterogeneous models, yet existing controllers are typically limited to one-shot routing: they select a model once and return its output directly. Such routing-only designs provide no mechanism to critique intermediate drafts or support iterative refinement.
To address this limitation, we propose a \emph{critique-and-routing controller} that casts multi-agent coordination as a sequential decision problem. At each turn, the controller evaluates the current draft, decides whether to stop or continue, and, if needed, selects the next agent for further refinement.
We formulate this process as a finite-horizon Markov Decision Process (MDP) with explicit agent-utilization constraints, design a composite reward for controller decisions across turns, and optimize the controller via policy gradients under a Lagrangian-relaxed objective.
Extensive experiments across multiple heterogeneous multi-agent systems and seven reasoning benchmarks show that our method consistently outperforms state-of-the-art baselines and substantially narrows the gap to the strongest agent, while using it for fewer than 25\% of total calls.
\end{abstract}

\section{Introduction}\label{sec:intro}


Multi-agent large language model (LLM) systems are becoming increasingly popular in practice~\citep{talebirad2023multi,li2023camel,wu2024autogen,li2024survey}. 
A representative deployment pattern is an \emph{agent pool} coordinated by a controller that selects among heterogeneous agents to balance quality and cost~\citep{ding2025best}. 
Most existing coordination mechanisms treat the controller primarily as a one-shot router, selecting a single model and returning the resulting response to the user~\citep{ding2024hybrid,lu2024routing}. 
However, many queries, especially those requiring sophisticated reasoning, are hard to solve in one turn: an initial draft may be incomplete or incorrect, and feedback can enable substantial refinement in subsequent turns~\citep{motwani2024malt}.
Yet one-shot routing-based coordination offers no such feedback loop: once an agent is selected, its response is returned without an opportunity for correction.

To mitigate the limitations of one-shot routing, recent work has explored multi-turn coordination from different perspectives. Prior work \cite{poon2026online} proposes a contextual-bandit framework where the router updates its selection strategy over rounds using user feedback. Router-R1~\citep{zhang2025router} trains an LLM router via reinforcement learning (RL) to perform multi-turn routing and aggregation, but primarily targets retrieval-style settings in which a query is decomposed into sub-queries and routed for information collection. Researchers in \cite{jin2025controlling} study cost-controllable coordination by conditioning the controller on externally provided budget modes, enabling different routing behavior under different budgets. Despite this progress, these approaches largely emphasize \emph{routing} and underexplore controller-to-agent \emph{interactive critique}, as well as explicit workload balancing across heterogeneous agents.

Based on these observations, we pose the following two questions:
\begin{enumerate}[leftmargin=*, itemsep=2pt, topsep=2pt, parsep=0pt, label=\emph{Q\arabic*:}]
\item Can we design a controller that jointly (i) routes to an appropriate agent, (ii) critiques and verifies intermediate drafts, and (iii) decides when to stop refining?
\item Can we optimize such a controller with a principled objective that jointly considers final-answer quality and system-level constraints such as balanced utilization across agents?
\end{enumerate}

\paragraph{Challenges.}


Answering these questions raises three key challenges. 
First, \emph{Joint routing and verification:} Unlike one-shot routers that only select a model, the controller must simultaneously assess response draft quality, decide whether to stop or continue refining, and select the next agent when refinement is needed. These decisions are tightly coupled: the verification shapes the next generation, and the stopping decision depends on the controller's assessment of the current draft.
\begin{wrapfigure}{r}{0.75\textwidth}
    \includegraphics[width=\linewidth]{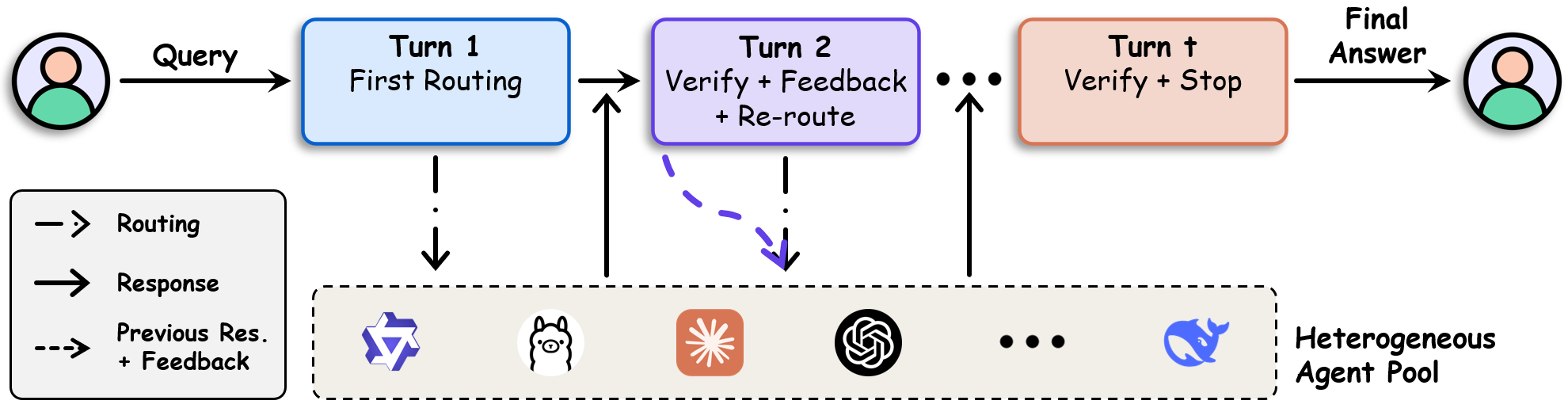}
    \caption{Controller interacts with a heterogeneous pool of agents over multiple rounds, evaluating the previous response from an agent and deciding whether to exit or refine the response.}
    \label{fig:system}
\end{wrapfigure}
Second, \emph{Trajectory-level credit assignment:} in a multi-turn interaction process, routing and verification decisions made at different turns of the same trajectory can contribute unequally to the final answer. However, commonly used outcome-based rewards blur these differences, obscuring which turns are primarily responsible for success or failure.
Third, \emph{Quality-utilization trade-off:} optimizing only for answer quality encourages overuse of the strongest, often the most expensive, agent, undermining the purpose of maintaining a heterogeneous agent pool. A practical controller should therefore be able to improve response quality through multi-turn interaction while balancing utilization across agents.


\subsection{Contributions}
To address these challenges, we study an iterative critique-and-routing controller and develop an RL-based framework for multi-turn coordination, which formulates multi-agent routing as sequential interaction control. The controller serves as both critic/verifier and router: at each turn, it evaluates the draft response produced by the agent selected in the previous turn and decides whether to stop or invoke another agent for further refinement. We formalize this as a finite-horizon Markov Decision Process (MDP) with explicit agent utilization constraints. Our main contributions are as follows:
\begin{itemize}[leftmargin=*, itemsep=2pt, topsep=2pt, parsep=0pt]
    \item \textbf{Sequential interaction formulation.} We propose a multi-turn controller with dual roles as critic and router, and formulate multi-agent coordination as a finite-horizon MDP for cumulative reward maximization. In this formulation, each controller action jointly specifies an evaluation of the previous response, a stop/continue decision, and the next agent to invoke. To prevent over-reliance on the strongest agent, we further impose explicit agent-utilization constraints.

\item \textbf{Reward and utilization-aware objective.} We introduce a composite rule-based reward that evaluates \emph{both} critique and routing decisions over trajectories. To facilitate optimization, we transform the constrained problem into an unconstrained one via Lagrangian relaxation, yielding a modified reward-maximization objective that can be optimized with standard policy gradient methods.

\item \textbf{Empirical evaluation.} Through extensive experiments on diverse heterogeneous multi-agent systems with different controllers, agents, and benchmarks, we demonstrate that our approach significantly outperforms baselines while maintaining comparable usage of the strongest agent.
\end{itemize}


\subsection{Related Work}\label{sec:related_work}

\paragraph{Multi-agent collaboration architectures.}
Prior work on multi-agent LLM systems can be broadly categorized into decentralized \emph{role-based} collaboration and centralized \emph{controller-agent} orchestration. In role-based systems, agents are assigned distinct roles and interact through predefined workflows~\citep{wu2024autogen,hong2023metagpt,qian2024chatdev}. A closely related variant is \emph{cascaded} specialization, in which different agents are invoked in a fixed sequence for generation, critique, and refinement, as in actor-critic style collaboration~\citep{estornell2024acc} and generate-verify-refine pipelines~\citep{motwani2024malt}. Recent work also considers query decomposition and routing, where complex tasks are decomposed into subproblems that are assigned to specialized agents~\citep{shao2025route,li2024agent,rasal2024navigating}. Another line of work studies centralized \emph{controller-agent} systems with a star-topology architecture, where a single controller coordinates a heterogeneous pool of experts and decides which agent to invoke for each query~\citep{chen2023frugalgpt,ding2024hybrid,ong2024routellm,ding2025best,jin2025controlling}. Our work focuses on this latter setting, and we discuss the most relevant controller design literature below.

\paragraph{Centralized controller-agent orchestration.}
Within controller-agent systems, coordination strategies range from lightweight classifier-based routing to reasoning-driven decision-making. 
Early approaches formulate coordination as a one-shot model-selection problem, using classification~\citep{lu2024routing,ding2025best}  or representation-learning-based routers \citep{mei2025omnirouter,chen2024routerdc} to dispatch each query to a single agent.
More recently, the authors in \cite{poon2026online} cast multi-agent coordination as a contextual bandit problem, accounting for multi-turn user-agent interaction and enabling the response quality to improve over time from observed human feedback. Router-R1~\citep{zhang2025router} and CoRL~\citep{jin2025controlling} further adopt reasoning-capable LLMs as controllers and apply RL to cultivate reasoning-aware routing policies. Despite this progress, existing methods largely treat coordination as isolated routing decisions, rather than as a sequential interaction process. Moreover, they typically lack explicit utilization constraints, leaving balanced and efficient use of heterogeneous agents underexplored.

\section{Problem Background}

\subsection{Controller Design for Multi-Agent Systems}
We consider a multi-agent LLM system with a heterogeneous agent pool $\mathcal{M} = \{\pi_{1}, \pi_{2}, \dots, \pi_{K}\}$ as shown in~\Cref{fig:system}, where agents differ in capability and cost. A controller coordinates this pool to respond to incoming queries. Unlike one-shot routing, where the controller selects a single model and directly returns its response without verification, we study a setting in which the controller coordinates the model pool over multiple turns by selecting an agent, evaluating/verifying the selected agent's response, and deciding whether to continue or terminate the interaction. The problem is therefore not merely to identify the single best model for a query in one turn, which is often difficult in practice, but to learn a controller policy that maximizes final task performance over multi-turn interactions while satisfying system-level usage constraints across agents. The controller should allocate calls across heterogeneous agents so as to improve answer quality through sequential interaction, without collapsing to always invoking the strongest or most expensive model.

\subsection{Aren't the Existing Solutions Good Enough?}
Although existing works have explored controller design in multi-agent systems from different perspectives, they still suffer from several limitations. Below, we discuss the two most relevant lines of work and describe their limitations in detail.

\textbf{Router-R1.}
Researchers in \cite{zhang2025router} introduce an RL-based controller for multi-agent LLM systems in information retrieval settings. Similar to Search-R1 \citep{jin2025search}, their approach decomposes a complex query into sub-queries and routes them to agents sequentially for information collection. However, this design is primarily tailored to retrieval-oriented scenarios, where queries are decomposable and the information returned by each agent is generally assumed to be correct and directly usable for subsequent reasoning, without explicit verification. Under such a design, incorrect or misleading intermediate outputs may propagate through the reasoning chain. In addition, Router-R1 is mainly optimized with outcome-level task rewards, which provide limited credit assignment in multi-turn trajectories: routing decisions made at different stages may contribute unequally to the final outcome, yet the reward signal does not clearly identify which intermediate actions should be rewarded.

\textbf{Cost-controllable routing.} CoRL~\citep{jin2025controlling} studies routing from the perspective of cost control by inducing different controller behaviors under different inference budgets. To achieve this, they manually design separate prompt templates for low-, medium-, and high-budget modes, and the routing policy adapts accordingly. However, the budget mode must be explicitly specified at inference time, rather than inferred autonomously from the query and interaction history. Furthermore, their framework focuses on routing under externally imposed budget settings and does not model iterative controller-agent interaction with critique, verification, and stopping decisions. Their objective also does not explicitly encourage balanced utilization across heterogeneous agents.

\section{RL-based Multi-Turn Coordination Methodology} 
\label{sec:methodology}
\vspace{-1mm}
In this section, we propose an \emph{iterative critique-and-routing controller} that models the agent coordination as a sequential decision problem. 
We formalize this as a finite-horizon MDP~\citep{puterman2014markov} with explicit load-balancing constraints and optimize the controller using a policy-gradient algorithm.

\subsection{Multi-Turn Coordination for Multi-Agent LLM Systems}\label{sec:multi_turn_dynamics}
Given a user query $\mathbf{q}$, the controller sequentially selects agents from the pool $\mathcal{M} = \{\pi_{1}, \pi_{2}, \dots, \pi_{K}\}$ and interacts with them to iteratively refine a response before producing the final output.
The coordination process unfolds over multiple turns as follows.
\begin{itemize}[leftmargin=*, itemsep=2pt, topsep=2pt, parsep=0pt]
\item \textbf{Turn 1 (initial routing).} For an incoming query $\mathbf{q}$, the controller, denoted as $\pi_{\boldsymbol{\theta}}$ and parameterized by $\boldsymbol{\theta}$, selects an agent to produce the initial response:
$$\mathbf{y}_1 = \pi_{\boldsymbol{\theta}}(\cdot \mid \mathbf{q}, \mathbf{c}), \quad m_1 = f(\mathbf{y}_1) \quad \longrightarrow \quad \mathbf{g}_1 = \pi_{m_1}(\cdot \mid \mathbf{q}),$$
where $\mathbf{c}$ denotes system prompt used for the initial routing and $f: \mathcal{Y} \to \mathcal{M}$ is a parsing function that extracts the discrete routing decision from controller's output text $\mathbf{y}$.
\item \textbf{Turn 2 (evaluation and re-routing).} After observing agent response $\mathbf{g}_1$, the controller evaluates its quality. If the response is deemed insufficient, the controller re-routes the query to another agent and proceeds to the next turn:
$$\mathbf{y}_2 = \pi_{\boldsymbol{\theta}}(\cdot \mid \mathbf{q}, \mathbf{c}^{\prime}, \mathbf{g}_1), \quad m_2 = f(\mathbf{y}_2) \quad \longrightarrow \quad \mathbf{g}_2 = \pi_{m_2}(\cdot \mid \mathbf{q}, \mathbf{g}_1, \mathbf{y}_2).$$
Here, $\mathbf{y}_2$ contains both a re-routing decision and an evaluation of $\mathbf{g}_1$, where the evaluation can be viewed as feedback that conditions the next generation.
For turns $t>1$, the prompt $\mathbf{c}^{\prime}$ differs from the first-turn prompt $\mathbf{c}$ by instructing the model to assess the draft response in addition to routing.
\item \textbf{Turn $t$ (general case).} For subsequent turns, the controller repeats the same pattern:
$$\mathbf{y}_t = \pi_{\boldsymbol{\theta}}(\cdot \mid \mathbf{q}, \mathbf{c}^{\prime}, \mathbf{g}_{t-1}), \quad m_t = f(\mathbf{y}_t) \quad \longrightarrow \quad \mathbf{g}_t = \pi_{m_t}(\cdot \mid \mathbf{q}, \mathbf{g}_{t-1}, \mathbf{y}_t).$$
The interaction continues until the controller chooses to terminate or the turn limit $T$ is reached.
\end{itemize}

In summary, the controller serves a dual role: (i) \emph{router}, selecting an agent from $\mathcal{M}$ and (ii) \emph{critic}, assessing response quality and generating feedback. We instantiate $\pi_{\boldsymbol{\theta}}$ as an LLM and design prompts $\mathbf{c}$ and $\mathbf{c}^{\prime}$, provided in \Cref{sec:appen:prompt}, to guide its behavior. To further strengthen these capabilities, we optimize the controller using RL, as detailed in the remainder of this section.


\subsection{Markov Decision Process Formulation}
We model the sequential interaction described in the previous subsection as a finite-horizon MDP defined by the tuple $(\mathcal{S}, \mathcal{A}, \mathcal{T}, \mathcal{R}, T)$. The state at turn $t > 1$ is $s_t = (\mathbf{q}, \mathbf{c}^{\prime}, \mathbf{g}_{t-1})$, comprising the user query, the system prompt, and the response generated in the previous turn. At the initial turn, $s_1 = (\mathbf{q}, \mathbf{c})$, where $\mathbf{c}$ is the initial system prompt. The action at turn $t$ is the full controller output $\mathbf{a}_t := \mathbf{y}_t = \pi_{\boldsymbol{\theta}}(\cdot \mid s_t)$, from which we parse a routing decision $m_t = f(\mathbf{y}_t)$. Upon executing action $\mathbf{a}_t$, the selected model $m_t$ produces a new response $\mathbf{g}_t = \pi_{m_t}(\cdot \mid \mathbf{q}, \mathbf{g}_{t-1}, \mathbf{y}_t)$, and the state transitions to $s_{t+1} = (\mathbf{q}, \mathbf{c}^{\prime}, \mathbf{g}_t)$. The episode terminates when the controller deems the current response draft sufficient, or when the maximum number of turns $T$ is reached.

\textbf{Reward.} Following the seminal work~\citep{guo2025deepseek}, we adopt a rule-based action reward. Each action $\mathbf{a}_t := \mathbf{y}_t$ jointly encodes routing and verification (i.e., evaluation). We define a binary verification reward $R_v(\mathbf{s}_t, \mathbf{a}_t)$ and a binary routing reward $R_r(\mathbf{s}_t, \mathbf{a}_t)$. The composite reward is:
\begin{equation}\label{reward_definition}
\begin{aligned}
r_t := R(\mathbf{s}_t, \mathbf{a}_t) &= 0.5\, R_r(\mathbf{s}_t, \mathbf{a}_t) + 0.5\, R_v(\mathbf{s}_t, \mathbf{a}_t), \\
R_r(\mathbf{s}_t, \mathbf{a}_t)&= \mathbf{1}\left(C_r(\mathbf{s}_t, \mathbf{a}_t)\right), \quad R_v(\mathbf{s}_t, \mathbf{a}_t)=\mathbf{1}\left(C_v(\mathbf{s}_t, \mathbf{a}_t)\right),
\end{aligned}
\end{equation}
where the reward mixing coefficient is set to $0.5$ to equally weight routing and verification.
$C_r(\mathbf{s}_t, \mathbf{a}_t)$ and $C_v(\mathbf{s}_t, \mathbf{a}_t)$ indicate whether the routing and verification decisions are correct, respectively. 
Notably, a routing decision is considered correct if the selected agent produces a correct solution to the problem. 
For boundary cases where one component is not applicable, we set the corresponding reward to zero (i.e., $R_v(\mathbf{s}_1,\mathbf{a}_1)=0$ and $R_r(\mathbf{s}_T,\mathbf{a}_T)=0$). 
Similar to~\citep{jin2025controlling,zhang2025router}, our current reward formulation targets verifiable domains where the correctness of both decisions is well-defined. Extending the framework to non-verifiable domains would require learned or LLM-based judges.


\textbf{Training Objective.} Under the above MDP formulation, executing the controller policy from the initial state $\mathbf{s}_1$ induced by the query $\mathbf{q}$ induces a trajectory
$\tau = (\mathbf{s}_1, \mathbf{a}_1, r_1,\; \mathbf{s}_2, \mathbf{a}_2, r_2,\; \ldots,\; \mathbf{s}_T, \mathbf{a}_T, r_T).$
Although some episodes terminate before turn $T$, we formulate the problem using a fixed horizon $T$ for notation convenience; after termination, the process accrues zero reward and no further agent invocation.
The controller is trained to maximize the expected discounted return:
\begin{equation}\label{equa:problem_formulation}
    \begin{aligned}
        \underset{\boldsymbol{\theta}}{\text{maximize}} ~~ & \mathbb{E}_{\mathbf{q}\sim \mathcal{D}}\, \mathbb{E}_{\mathbf{a}_t \sim \pi_{\boldsymbol{\theta}}(\cdot \mid \mathbf{s}_t)}\!\left[\sum_{t=1}^{T} \gamma^{t-1} R(\mathbf{s}_t, \mathbf{a}_t)\right], \\
        \text{subject to} ~~ & \sum_{t=1}^T \mathbb{E}\bigl[\mathbf{1}(f(\mathbf{a}_t)=k)\bigr] \leq \rho_k \sum_{k^{\prime}=1}^K\sum_{t=1}^T \mathbb{E}\bigl[\mathbf{1}(f(\mathbf{a}_t)=k^{\prime})\bigr], \quad \forall k,
    \end{aligned}
\end{equation}
where $\gamma \in (0, 1]$ is a discount factor, query $\mathbf{q}$ is encoded in $\mathbf{s}_t$, 
$\mathbf{1}(f(\mathbf{a}_t) = k)$ indicates whether agent $k$ is invoked at turn $t$, $\sum_{t=1}^T \mathbb{E}[\mathbf{1}(f(\mathbf{a}_t)=k)]$ is the expected invocation count of agent $k$, and $\sum_{k=1}^K\sum_{t=1}^T \mathbb{E}[\mathbf{1}(f(\mathbf{a}_t)=k)]$ is the expected total invocation count across all agents. The constraint, configured by $\rho_k$, limits the usage of each agent and promotes load balance across the agent pool.

\subsection{Lagrangian Relaxation}\label{sec:lagrangian_relaxation_derivation}

Problem~\eqref{equa:problem_formulation} is a constrained reward maximization that is difficult to optimize directly using existing policy gradient methods. We therefore reformulate it via Lagrangian relaxation~\citep{schulman2017proximal}. The resulting unconstrained objective is given in Proposition~\ref{Proposition_lagrangian_relaxation}. The proof is provided in Appendix~\ref{sec:appen:proof_proposition_relaxation}.
\begin{proposition}\label{Proposition_lagrangian_relaxation}
Suppose that the discount factor $\gamma = 1$. Define the modified composite reward and routing reward as:
\begin{equation}\label{modified_reward}
\resizebox{0.93\linewidth}{!}{%
$R^{\prime}(\mathbf{a}_t, \mathbf{s}_t) := 0.5 R^{\prime}_r(\mathbf{a}_t, \mathbf{s}_t) + 0.5 \, R_v(\mathbf{s}_t, \mathbf{a}_t), ~~  R^{\prime}_r(\mathbf{a}_t, \mathbf{s}_t) := R_r(\mathbf{s}_t, \mathbf{a}_t) - \sum_{k=1}^K 2 \xi_k \cdot \mathbf{1}(f(\mathbf{a}_t)\!=\!k)$,
}
\end{equation}
where $\xi_k \geq 0$ is the tunable Lagrange multiplier associated with the usage constraint on agent $k$.
Problem~\eqref{equa:problem_formulation} can be optimized through the following unconstrained modified-reward maximization:
\begin{equation}\label{equa:problem_formulation_relaxed_4}
    \begin{aligned}
        \underset{\boldsymbol{\theta}}{\text{maximize}} ~~ \mathbb{E}_{\mathbf{q}\sim \mathcal{D}}\, \mathbb{E}_{\mathbf{a}_t \sim \pi_{\boldsymbol{\theta}}(\cdot \mid \mathbf{s}_t)}\!\left[\sum_{t=1}^{T} \gamma^{t-1} R^{\prime}(\mathbf{a}_t, \mathbf{s}_t) \right]. 
    \end{aligned}
\end{equation}
\end{proposition}

\begin{remark}[Interpretation of the Modified Routing Reward]
The modified routing reward $R^{\prime}_r(\mathbf{a}_t, \mathbf{s}_t)$ penalizes over-reliance on individual agents by subtracting a usage cost proportional to $\xi_k$ whenever agent $k$ is selected. When the invocation frequency of agent $k$ exceeds its budget, increasing $\xi_k$ reduces its effective routing reward, discouraging the controller from repeatedly routing to that agent. This mechanism provides a configurable knob to balance workload across the pool without modifying the underlying optimization algorithm.
\end{remark}
\begin{remark}[Justification of the Relaxation]
The constraints in \eqref{equa:problem_formulation} are designed to encourage load-balanced utilization and are not necessarily hard requirements at deployment time. While the relaxation in \eqref{equa:problem_formulation_relaxed_4} is not strictly equivalent to the original constrained formulation, it yields a theoretically grounded surrogate objective that is well suited to standard policy-gradient algorithms.
\end{remark}

\begin{figure}[t]
    \centering
    \includegraphics[width=\textwidth]{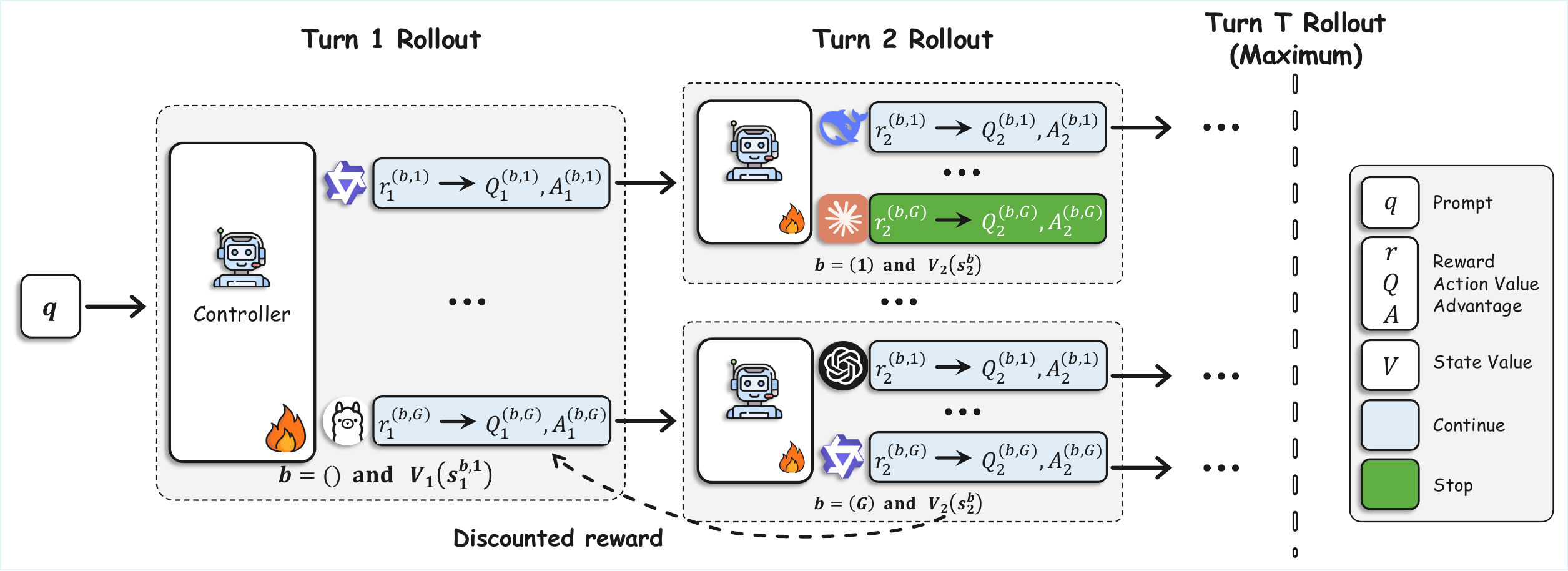}
        \caption{Multi-turn rollout tree for GRPO training of the critique-and-routing controller. From each state $\mathbf{s}_t^{b}$ with prefix $b$ (starting from the root prefix $()$), the controller samples a group of actions $\mathbf{a}_t^{(b,i)}$, yielding a group of new prefixes $\{(b,i)\}_{i=1}^{G}$. Non-terminal actions invoke an agent to produce the next state $\mathbf{s}_{t+1}^{(b,i)}$, while each sampled action receives a modified reward that incorporates routing quality, verification quality, and an agent-utilization penalty.}
    \label{fig:multi_turn_reward}
    \vspace{-3mm}
\end{figure}


\subsection{Policy Gradient Methods for Controller Training}

Problem \eqref{equa:problem_formulation_relaxed_4} can be optimized with established policy-gradient algorithms. 
In this work, we adopt a commonly used variant of policy-gradient methods in LLM, \emph{Group Relative Policy Optimization} (GRPO)~\citep{shao2024deepseekmath}, which samples multiple rollouts per query and optimizes the policy using group-relative advantage. The detailed justification of tailoring GRPO to Problem \eqref{equa:problem_formulation_relaxed_4} is provided in~\Cref{sec:appen:grpo}. In this section, we convert the multi-turn discounted reward into group-relative advantage, a key component of GRPO.

Following the interaction procedure in~\Cref{sec:multi_turn_dynamics}, GRPO samples multiple rollouts for each query at each turn. To distinguish these trajectories, we introduce an additional index $b$, and denote by $\mathbf{s}_t^b$ the state at turn $t$ along rollout $b$ (see ~\Cref{fig:multi_turn_reward}). This notation is used only to facilitate the recursive computation of discounted returns under group sampling. From each state $\mathbf{s}_t^b$, we sample $G$ actions $\{\mathbf{a}_t^{(b,i)}\}_{i=1}^{G}$ (i.e., responses) from the controller $\pi_{\boldsymbol{\theta}}(\cdot \mid \mathbf{s}_t^{b})$. Each sampled action yields an immediate modified action reward via \eqref{modified_reward},
\[
r_t^{(b,i)} = R'(\mathbf{s}_t^{b}, \mathbf{a}_t^{(b,i)}),
\]
and induces a next state $\mathbf{s}_{t+1}^{(b,i)}$. The discounted return defined in Problem \eqref{equa:problem_formulation_relaxed_4} at state $\mathbf{s}_t^{b}$ is thus
\vspace{-1mm}
\begin{equation}
\hat{V}_t(\mathbf{s}_t^{b}) =
\frac{1}{G}\sum_{i=1}^{G}
\left[
r_t^{(b,i)} + \gamma \hat{V}_{t+1}(\mathbf{s}_{t+1}^{(b,i)})
\right],
\end{equation}
\vspace{-1mm}
with terminal condition $\hat{V}_{T+1}(\cdot) = 0$. Using this discounted return, we define the group relative advantage for action $\mathbf{a}_t^{(b,i)}$ as
\begin{equation}
A_t^{(b,i)} =
\hat{Q}_t^{(b,i)} - \mathrm{mean}\!\left(\{\hat{Q}_t^{(b,j)}\}_{j=1}^{G}\right),
\quad
\hat{Q}_t^{(b,i)} =
r_t^{(b,i)} + \gamma \hat{V}_{t+1}\!\left(\mathbf{s}_{t+1}^{(b,i)}\right).
\end{equation}
With this advantage estimate, we then optimize the controller policy using the standard GRPO algorithm, which is detailed in~\Cref{sec:appen:grpo_algorithm}.

\begin{table}[t]
    \footnotesize
    \centering
    \setlength{\tabcolsep}{4.5pt}
    \rowcolors{1}{white}{gray!10}
    \caption{\textbf{Accuracy comparison for two heterogeneous multi-agent systems with different controllers and agent pools.} 
    We abbreviate MATH-lighteval as M-LEval, Omni-MATH as Omni, MinervaMath as Minerva, and AGI-Eval-MATH as AGI-M.
    Overall, our method achieves the highest average accuracy among baselines in both systems under \emph{usage constraints} on Agents 1 and 2.}
    \label{tab:exp_main}
    \vspace{1mm}
    \begin{tabular}{lcccccccc}
        \toprule
        \rowcolor{white}
        \textbf{Methods} & \textbf{M-LEval} & \textbf{MATH-500} & \textbf{Omni} & \textbf{Minerva} & \textbf{AGI-M} & \textbf{AMC23} & \makecell[c]{\textbf{AIME24}} & \textbf{Avg.} \\
        \cmidrule{2-9}
        \rowcolor{white}
        \multicolumn{9}{c}{\textbf{Controller: Qwen2.5-7B-Base}} \\
        \rowcolor{white}
        \multicolumn{9}{p{\dimexpr\linewidth-2\tabcolsep\relax}}{\textbf{Agent 1}: Qwen3-30B-A3B-Instruct; \textbf{Agent 2}: Qwen2.5-7B-Instruct; \textbf{Agent 3}: Qwen2.5-1.5B-Instruct.} \\
        \rowcolor{white}
        \midrule
        \rowcolor{white}
        Agent 1 & 0.956 & 0.938 & 0.551 & 0.828 & 0.933 & 0.975 & 0.833 & 0.859 \\
        Agent 2 & 0.757 & 0.750 & 0.254 & 0.524 & 0.734 & 0.500 & 0.100 & 0.517 \\
        Agent 3 & 0.554 & 0.542 & 0.157 & 0.273 & 0.547 & 0.275 & 0.067 & 0.345 \\
        \midrule
        Controller V1 & 0.626 & 0.596 & 0.205 & 0.247 & 0.597 & 0.450 & 0.100 & 0.403 \\
        Controller V2 & 0.749 & 0.760 & 0.246 & 0.423 & 0.724 & 0.600 & 0.200 & 0.529 \\
        Random Router & 0.756 & 0.744 & 0.304 & 0.537 & 0.737 & 0.575 & 0.275 & 0.561 \\
        RoBERTa Router  & 0.763 & 0.750 & 0.436 & 0.417 & 0.729 & 0.550 & 0.667 & 0.616 \\
        RouterDC & 0.784 & 0.764 & 0.445 & 0.419 & 0.736 & 0.600 & 0.275 & 0.575 \\
        Router-R1  & 0.776 & 0.778 & 0.495 & 0.432 & 0.763 & 0.825 & 0.767 & 0.691 \\
        \midrule
        \rowcolor{row-highlight}\textbf{Ours}  & \textbf{0.901}  & \textbf{0.888} & \textbf{0.534} & \textbf{0.678} & \textbf{0.879} & \textbf{0.925} & \textbf{0.833} & \textbf{0.805} \\
        \midrule
        \rowcolor{white}
        \multicolumn{9}{c}{\textbf{Controller: Qwen3-4B-Base}} \\
        \rowcolor{white}
        \multicolumn{9}{p{\dimexpr\linewidth-2\tabcolsep\relax}}{\textbf{Agent 1}: Qwen3-30B-A3B-Instruct; \textbf{Agent 2}: Ministral-3-8B-Instruct; \textbf{Agent 3}: Llama-3-2-1B-Instruct.} \\
        \midrule
        Agent 1 & 0.956 & 0.938 & 0.551 & 0.828 & 0.933 & 0.975 & 0.833 & 0.859 \\
        Agent 2 & 0.724 & 0.720 & 0.180 & 0.599 & 0.681 & 0.400 & 0.100 & 0.486 \\
        Agent 3 & 0.275 & 0.260 & 0.067 & 0.088 & 0.265 & 0.200 & 0.000 & 0.165 \\
        \midrule
        Controller V1 & 0.761 & 0.738 & 0.255 & 0.460 & 0.736 & 0.550 & 0.133 & 0.519 \\
        Controller V2 & 0.883 & 0.804 & 0.328 & 0.419 & 0.809 & 0.775 & 0.333 & 0.623 \\
        Random Router & 0.670 & 0.659 & 0.244 & 0.528 & 0.640 & 0.500 & 0.267 & 0.501 \\
        RoBERTa Router  & 0.659 & 0.662 & 0.458 & 0.376 & 0.628 & 0.775 & 0.833 & 0.627 \\
        RouterDC & 0.681 & 0.674 & 0.463 & 0.386 & 0.654 & 0.600 & 0.467 & 0.561 \\
        Router-R1  & 0.686 & 0.682 & 0.492 & 0.405 & 0.666 & 0.925 & 0.800 & 0.665 \\
        \midrule
        \rowcolor{row-highlight}\textbf{Ours}  & \textbf{0.915}  & \textbf{0.896} & \textbf{0.541} & \textbf{0.718} & \textbf{0.881} & \textbf{0.975} & \textbf{0.833} & \textbf{0.823} \\
       \bottomrule
    \end{tabular}
    \vspace{-6mm}
\end{table}

\vspace{-1mm}
\section{Experiments}\label{sec:main_experiments}
\vspace{-1mm}
\textbf{Datasets.} We train the controller on the training split of MATH-lighteval~\citep{hendrycks2measuring} and evaluate it on the corresponding test split and six additional math benchmarks: MATH500~\citep{hendrycks2measuring},
Omni-MATH~\citep{gao2024omni}, MinervaMath~\citep{li2024numinamath},
AGI-Eval-MATH~\citep{zhong2024agieval}, AMC23~\citep{lewkowycz2022solving},
and AIME24. 
Together, these benchmarks provide a comprehensive testbed for evaluating effectiveness and generalization of our method.

\textbf{Model Pools.} We consider two heterogeneous multi-agent systems with different controllers and agent pools. In System 1, the controller is initialized with Qwen2.5-7B-Base,
and the candidate agents are Qwen3-30B-A3B-Instruct, Qwen2.5-7B-Instruct, and
Qwen2.5-1.5B-Instruct~\citep{hui2024qwen2,yang2025qwen3}. In System 2,
we use Qwen3-4B-Base as the controller, and the candidate agents are Qwen3-30B-A3B-Instruct,
Ministral-3-8B-Instruct-2512, and Llama-3-2-1B-Instruct~\citep{hui2024qwen2,yang2025qwen3}.
We also study additional controller choices; the corresponding results are reported
in Appendix~\ref{sec:appen:other_controller}.

\textbf{Baselines.} We compare our method with the following baselines.
\begin{itemize}[leftmargin=*, itemsep=2pt, topsep=2pt, parsep=0pt]
    \item Router-R1: a state-of-the-art RL-based reasoning router~\citep{zhang2025router}. Since
    the query is not decomposable in our setting, the controller routes the full problem
    to a single agent.
    \item RouterDC~\citep{chen2024routerdc}: a contrastive-learning-based router that maps queries and agents into a shared embedding space and selects agents based on embedding similarity.
    \item RoBERTa Router: trains a RoBERTa-based classifier with supervised learning to predict the best agent for a
    given query~\citep{ding2024hybrid,ong2024routellm}.
    \item Random Router: randomly selects an agent without any learned routing policy.
    \item Controller V1: uses the controller itself as the solver and trains it with
    RL-based post-training (GRPO) using an outcome-based reward~\citep{guo2025deepseek}.
    \item Controller V2: trains the controller with our method and directly uses it as
    a standalone solver at inference time.
\end{itemize}

\textbf{Resource Constraints.} To avoid over-reliance on the strongest agent and to
encourage balanced utilization, we impose a usage budget on that agent during the training. Specifically,
we set its usage constraint to $\rho_1 = 0.25$, meaning that it can be selected in at
most 25\% of the whole invocations. In addition, $\rho_2$ and $\rho_3$ are set to be $0.5$ and $1$, respectively. These constraints can be satisfied by setting appropriate Lagrange multipliers $\{\xi_k\}_{k=1}^3$ (see~\Cref{tab:hyperparameters}). We discuss the choice of $\rho_1$ and $\rho_2$ and further study the sensitivity of $\xi_k$ in~\Cref{sec:appen:lagrangian_sensitivity}.


\textbf{Other Details.} Our experiments are implemented using the verl framework~\citep{sheng2025hybridflow}. We conduct all experiments on a cluster with 4 NVIDIA
H100 GPUs, each with 80 GB of memory. Additional implementation details and
hyperparameter settings are provided in Appendix~\ref{sec:appen:hyperparameter}.

\begin{figure}[t]
    \centering
    \begin{subfigure}[b]{0.32\textwidth}
        \centering
        \includegraphics[width=\textwidth]{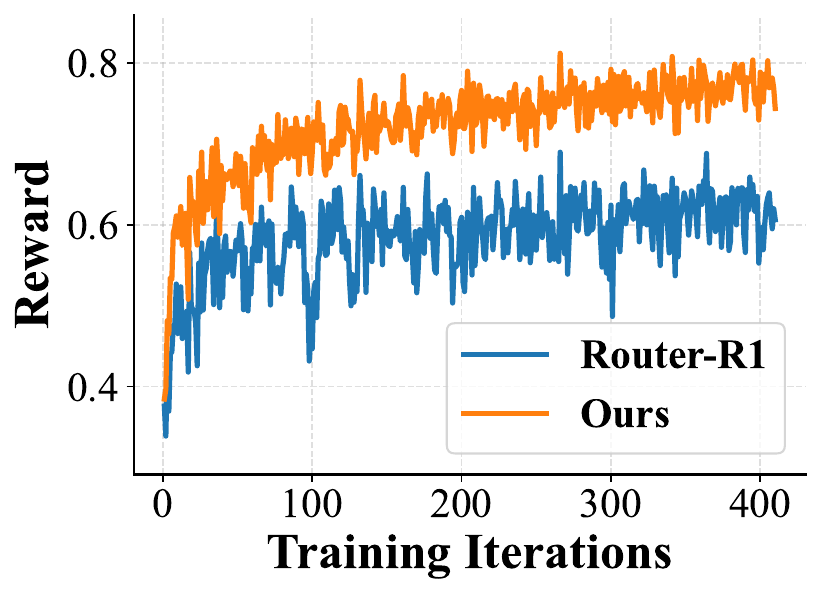}
        \caption{Reward in System 1}
        \label{fig:reward_system1}
    \end{subfigure}\hfill
    \begin{subfigure}[b]{0.32\textwidth}
        \centering
        \includegraphics[width=\textwidth]{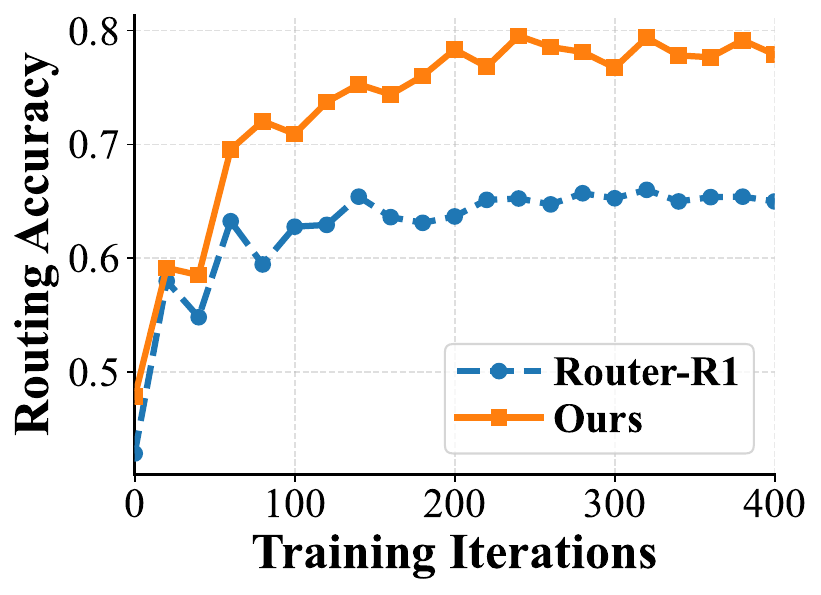}
        \caption{Accuracy in System 1}
        \label{fig:routing_accuracy_system1}
    \end{subfigure}\hfill
    \begin{subfigure}[b]{0.32\textwidth}
        \centering
        \includegraphics[width=\textwidth]{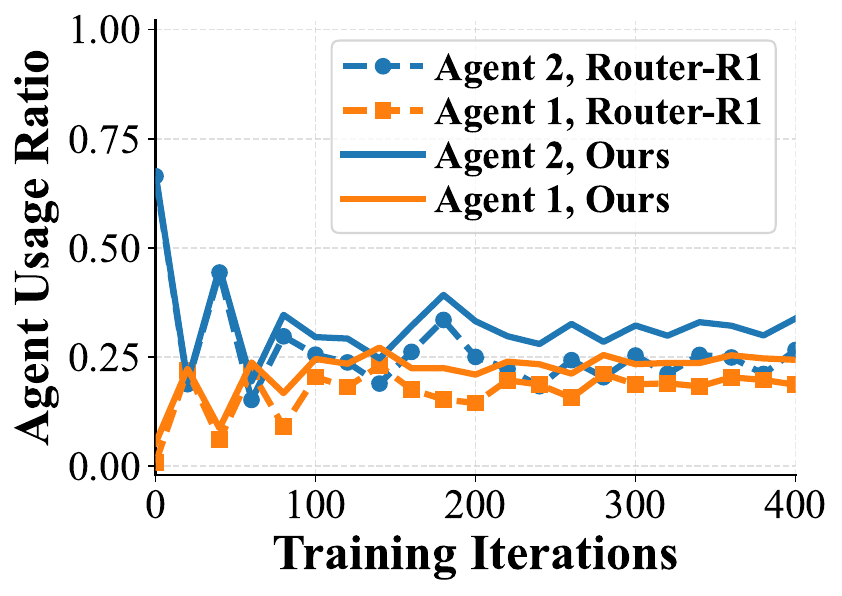}
        \caption{Agent's usage in System 1}
        \label{fig:model_usage_system1}
    \end{subfigure}

    \vspace{0.5em}

    \begin{subfigure}[b]{0.32\textwidth}
        \centering
        \includegraphics[width=\textwidth]{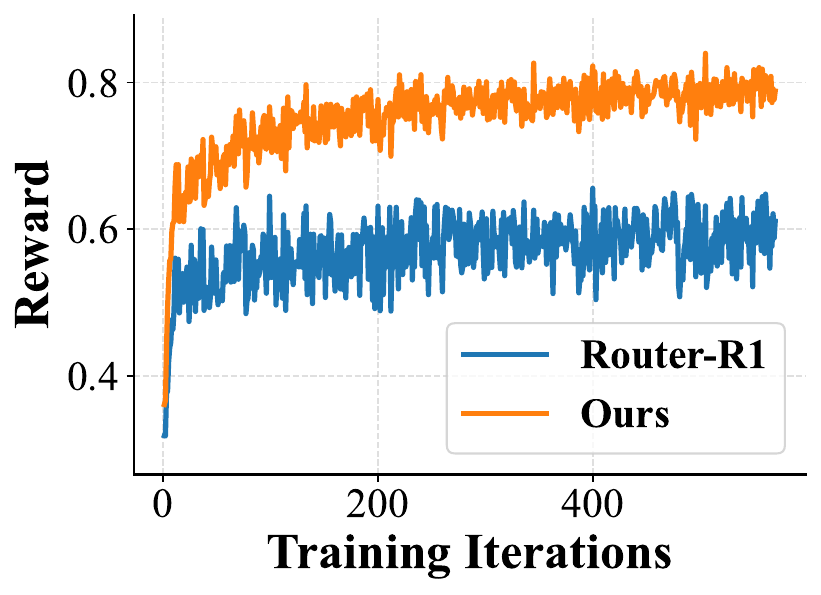}
        \caption{Reward in System 2}
        \label{fig:reward_system2}
    \end{subfigure}\hfill
    \begin{subfigure}[b]{0.32\textwidth}
        \centering
        \includegraphics[width=\textwidth]{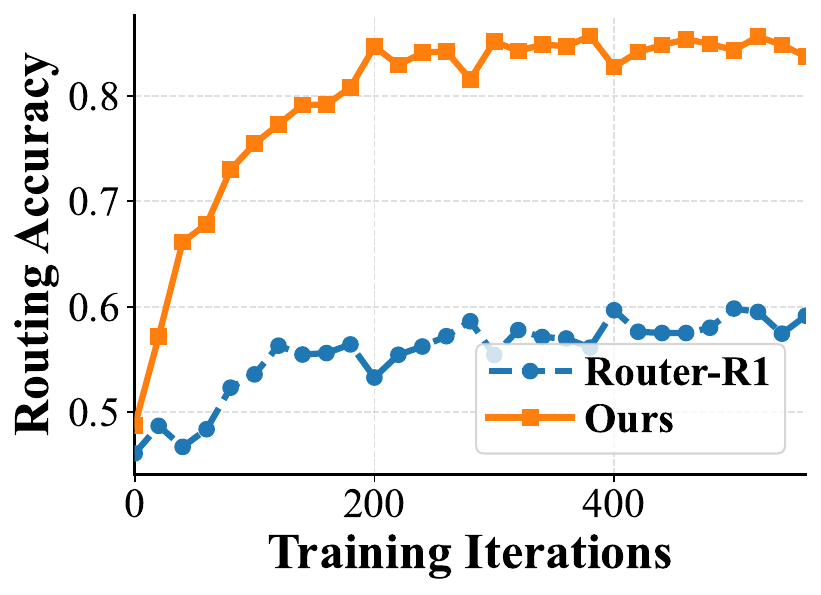}
        \caption{Accuracy in System 2}
        \label{fig:routing_accuracy_system2}
    \end{subfigure}\hfill
    \begin{subfigure}[b]{0.32\textwidth}
        \centering
        \includegraphics[width=\textwidth]{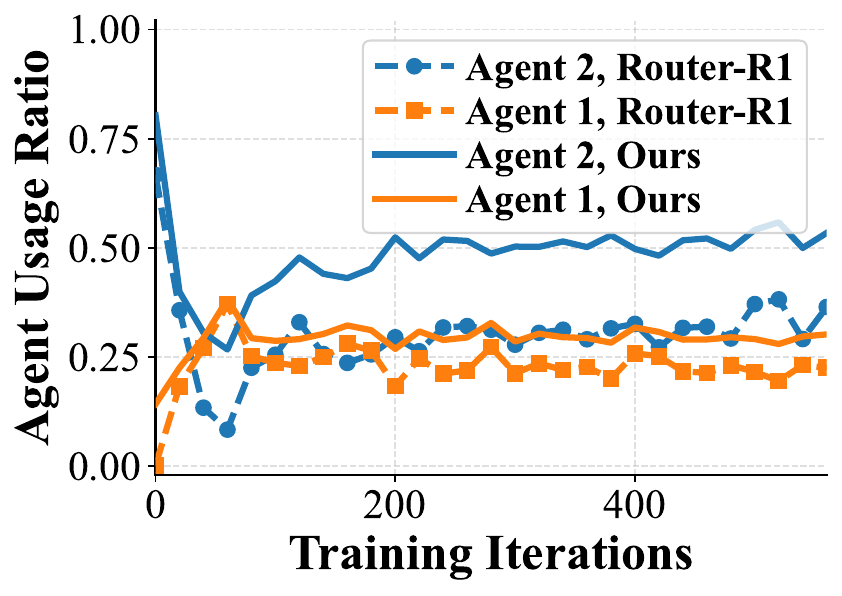}
        \caption{Agent's usage in System 2}
        \label{fig:model_usage_system2}
    \end{subfigure}
    \caption{\textbf{Training dynamics for two multi-agent systems with different controller and agent-pool configurations.}
	System~1 uses Qwen2.5-7B-Base as the controller with agents Qwen3-30B-A3B-Instruct (Agent 1), Qwen2.5-7B-Instruct (Agent 2), and Qwen2.5-1.5B-Instruct. System~2 uses Qwen3-4B-Base as the controller with agents Qwen3-30B-A3B-Instruct (Agent 1), Ministral-3-8B-Instruct (Agent 2), and Llama-3-2-1B-Instruct. 
    Compared with Router-R1, our method significantly improves reward and accuracy with slightly higher usage on the strongest agent.}
    \label{fig:llm-training-dynamics}
    \vspace{-5mm}
\end{figure}

\vspace{-2mm}
\subsection{Main Results}

In Table~\ref{tab:exp_main}, we compare the accuracy of our method with several baselines across seven math benchmarks under two heterogeneous multi-agent systems. 
In System 1, where the controller is Qwen2.5-7B-Base and the agents have sizes 30B, 7B, and 1.5B, our method achieves an average accuracy of 80.5\%. 
It outperforms the strongest routing baseline, Router-R1 (69.1\%), by 11.4 percentage points. 
The advantage is consistent across these benchmarks and is particularly pronounced on challenging benchmarks such as AGI-Eval-MATH and MinervaMath, highlighting the effectiveness of our approach in coordinating multiple agents for complex reasoning tasks.
In System 2, with Qwen3-4B-Base as the controller and agents of sizes 30B, 8B, and 1B, our method further improves the average accuracy to 82.3\%, surpassing Router-R1 (66.5\%) by 15.8 percentage points. 
The strong results across both systems indicate that our method with \emph{explicit multi-turn interaction} generalizes well across different controller capacities and agent compositions. 
Furthermore, the comparison with baselines Controller V1 and Controller V2 suggests that the gains are related not only to controller strength, but also to more effective coordination behavior.

In Figure~\ref{fig:llm-training-dynamics}, we present the training dynamics of the controllers in the two multi-agent systems under Router-R1 and our method, in terms of reward, routing accuracy, and agent usage ratio throughout training. 
The agent usage ratio is defined as the number of times an agent is invoked, normalized by the dataset size.
Other baselines are excluded since their training dynamics are not directly comparable.

\emph{Reward trajectories:} From~\Cref{fig:reward_system1,fig:reward_system2}, we see that our method achieves rapid reward improvement across both systems, reaching stable high-reward plateaus within the first 100 steps, with final rewards substantially higher than Router-R1. 
This demonstrates that the controller learns both effective routing and verification behaviors over multiple turns that lead to a higher reward.

\emph{Routing accuracy trends:} The routing accuracy curves in~\Cref{fig:routing_accuracy_system1,fig:routing_accuracy_system2} show that our method learns a substantially more accurate routing policy than Router-R1 in both systems. 
This improvement suggests that the \emph{multi-turn interaction} mechanism provides additional opportunities for the controller to refine or correct earlier routing decisions. 
It also confirms that the proposed method achieves meaningful gains in routing policy optimization, which is a critical factor underlying the overall accuracy improvement shown in~\Cref{tab:exp_main}.

\emph{Agent utilization patterns:} \Cref{fig:model_usage_system1,fig:model_usage_system2} report the usage ratios of the strongest and intermediate agents. Overall, our method leads to only a modest increase in strongest-agent usage relative to Router-R1, suggesting that the performance gains are not simply due to over-reliance on the strongest agent. 
This can be attributed to the hierarchical reward design that encourages more effective use of stronger agents while still respecting the usage constraint.
In System~2, our method also shows higher utilization of the intermediate-sized agent, which is consistent with the explicit multi-turn interaction mechanism in our framework. 
In summary, these results indicate that our method achieves a favorable balance between performance and agent utilization.

\subsection{Ablation Studies}
To assess the contribution of each component in our framework, we conduct ablation studies on the number of interaction turns and the Lagrange multipliers.

\begin{wrapfigure}{r}{0.68\textwidth}
    \centering
    \begin{minipage}{0.32\textwidth}
        \centering
        \includegraphics[width=\textwidth]{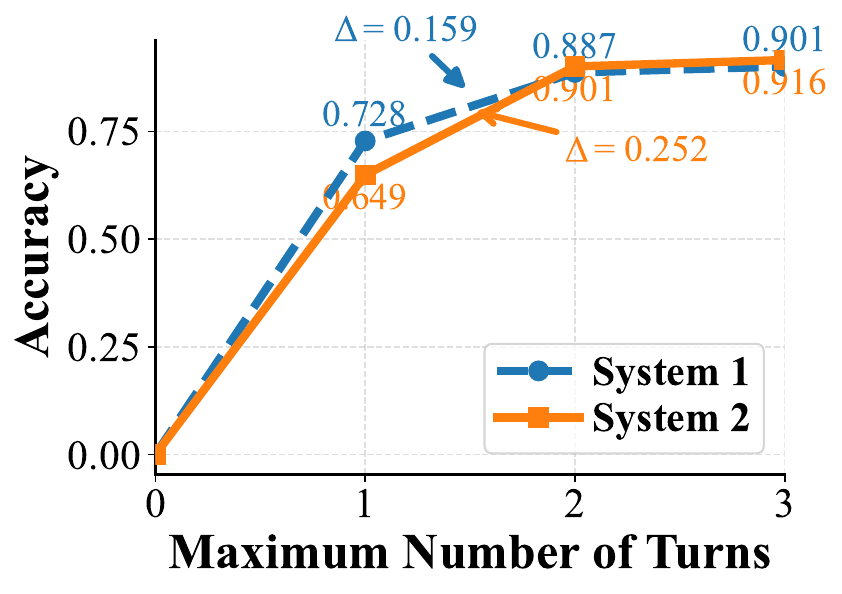}
        \caption{Accuracy under different maximum turn budgets}
        \label{fig:accuracy_turns}
    \end{minipage}
    \hspace{2mm}
    \begin{minipage}{0.32\textwidth}
        \centering
        \includegraphics[width=\textwidth]{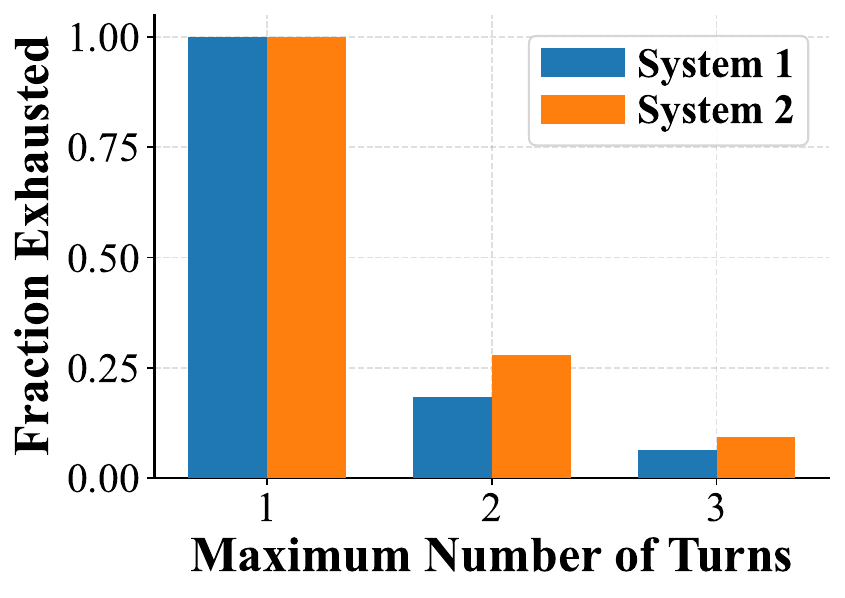}
        \caption{Percentage of prompts exhausting the turn budget.}
        \label{fig:fraction_exhausted_turns}
    \end{minipage}
    \vspace{-3mm}
\end{wrapfigure}

\textbf{Impact of the Number of Turns.}
We vary the maximum number of interaction turns $T$ at inference time and report the accuracy on MATH-lighteval benchmark in~\Cref{fig:accuracy_turns} for both Systems 1 and 2. Notably, the maximum number of interaction turns used during training does not need to match that used during inference. 
Multi-turn training equips the controller with the ability to interact over multiple rounds, without requiring an exact match between training and inference horizons.
As shown in~\Cref{fig:accuracy_turns}, the accuracy increases significantly when the maximum number of turns is increased from 1 to 2, which further validates the effectiveness of multi-turn interaction in boosting the accuracy. We also observe that the performance gains saturate as the number of turns continues to increase, suggesting that a small turn budget is already sufficient for most prompts. 
This is further supported by~\Cref{fig:fraction_exhausted_turns}, which shows that only a small fraction of prompts exhaust the turn budget under larger maximum-turn settings.

\textbf{Impact of the Lagrange Multipliers.}
We introduce Lagrange multipliers $\xi_k$ to discourage over-reliance on the $k$-th agent. To evaluate their effect, we compare system performance with and without these multipliers. For the three-agent systems considered here, the setting with Lagrange multipliers uses $\xi_1=0.25$, $\xi_2=0.125$, and $\xi_3=0$, while the setting without them sets $\xi_1=\xi_2=\xi_3=0$ in~\Cref{modified_reward}. We observe that removing the Lagrange multipliers results in uncontrolled agent usage, where all the queries are routed to the strongest agent. We report the overall accuracy in~\Cref{tab:lagrangian} under the usage constraints $\rho_1=0.25$ and $\rho_2=0.5$. During evaluation, once an agent's budget is exhausted, incoming requests are automatically routed to the next-best available agent for both our method and the baselines. \Cref{tab:lagrangian} demonstrates that the Lagrangian penalty effectively enables the system to achieve higher accuracy under constrained agent usage.

\begin{table}[t]
    \footnotesize
    \centering
    \setlength{\tabcolsep}{4.5pt}
    \rowcolors{1}{white}{gray!10}
    \caption{\textbf{Comparison on seven math benchmarks for two heterogeneous multi-agent systems with and without Lagrangian multipliers.} 
    We abbreviate MATH-lighteval as M-LEval, Omni-MATH as Omni, MinervaMath as Minerva, and AGI-Eval-MATH as AGI-M.}
    \label{tab:lagrangian}
    \vspace{1mm}
    \begin{tabular}{lcccccccc}
        \toprule
        \rowcolor{white}
        \textbf{Methods} & \textbf{M-LEval} & \textbf{MATH-500} & \textbf{Omni} & \textbf{Minerva} & \textbf{AGI-M} & \textbf{AMC23} & \makecell[c]{\textbf{AIME24}} & \textbf{Avg.} \\
        \cmidrule{2-9}
        \rowcolor{white}
        \multicolumn{9}{c}{\textbf{System 1}} \\
        \rowcolor{white}
        \midrule
        \rowcolor{white}
        W/O Lagrangian & 0.756 & 0.746 & 0.304 & 0.537 & 0.737 & 0.575 & 0.300 & 0.565 \\
        \midrule
        \rowcolor{row-highlight}\textbf{W Lagrangian}  & \textbf{0.901}  & \textbf{0.888} & \textbf{0.534} & \textbf{0.678} & \textbf{0.879} & \textbf{0.925} & \textbf{0.833} & \textbf{0.805} \\
        \midrule
        \rowcolor{white}
        \multicolumn{9}{c}{\textbf{System 2}} \\
        \midrule
        W/O Lagrangian & 0.670 & 0.660 & 0.245 & 0.529 & 0.640 & 0.500 & 0.267 & 0.501 \\
        \midrule
        \rowcolor{row-highlight}\textbf{W Lagrangian}  & \textbf{0.915}  & \textbf{0.896} & \textbf{0.541} & \textbf{0.718} & \textbf{0.881} & \textbf{0.975} & \textbf{0.833} & \textbf{0.823} \\
       \bottomrule
    \end{tabular}
    \vspace{-3mm}
\end{table}

\vspace{-2mm}
\section{Conclusion and Limitations}\label{sec:conclusion}

We studied controller design for multi-turn coordination in heterogeneous multi-agent LLM systems. We proposed an iterative critique-and-routing controller that evaluates intermediate drafts, decides whether to stop or continue, and selects the next agent for further refinement when needed. We formulated this process as a finite-horizon MDP with explicit agent-utilization constraints, designed a composite reward for decisions across turns, and optimized the controller via policy gradients under a Lagrangian-relaxed objective. Extensive experiments across multiple heterogeneous multi-agent systems and seven reasoning benchmarks demonstrate that our method consistently outperforms strong routing baselines and substantially narrows the gap to the strongest agent while using it for fewer than 25\% of total calls. Overall, these results highlight the value of iterative refinement in multi-agent LLM systems.
A limitation of our work is that experiments primarily focus on math reasoning benchmarks, where routing and verification rewards are relatively straightforward to define. In addition, the multi-turn design introduces extra inference latency and computational overhead compared to one-shot routing, reflecting a trade-off between performance and efficiency.

\bibliography{ref.bib}
\bibliographystyle{plain}

\newpage
\appendix
\clearpage

\begin{center}
    {\bf\Large Appendix}
\end{center}

\startcontents[sections]
\printcontents[sections]{l}{1}{\setcounter{tocdepth}{4}}

\clearpage
\section{Further Experiments}
\subsection{Other Controller Choices}\label{sec:appen:other_controller}

Table~\ref{tab:exp_other_controller} reports an additional study under the same experimental setup as~\Cref{tab:exp_main}, where we replace the controller backbone with alternative models while keeping the agent pools, benchmarks, and baselines unchanged. This allows us to isolate the effect of the controller and evaluate whether our training method generalizes across different controller capacities.
We observe that our method consistently achieves the best performance across all benchmarks and both system configurations, regardless of the underlying controller. In particular, under the Qwen3-4B-Base controller, our method attains an average accuracy of 80.3\%, outperforming the strongest baseline (Router-R1, 72.4\%) by a substantial margin. Similarly, when using Qwen3-8B-Base as the controller, our method further improves the average accuracy to 82.3\%, exceeding Router-R1 (64.9\%) by 17.4 percentage points.
Overall, these results, together with~\Cref{tab:exp_main}, suggest that the proposed method exhibits robustness to the choice of controller across a range of settings, supporting its applicability in heterogeneous multi-agent systems.


\begin{table}[t]
    \footnotesize
    \centering
    \setlength{\tabcolsep}{4.5pt}
    \rowcolors{1}{white}{gray!10}
    \caption{Accuracy comparison on seven math datasets for two multi-agent systems using alternative controller models under \emph{usage constraints} on Agents 1 and 2. We abbreviate MATH-lighteval as M-LEval, Omni-MATH as Omni, MinervaMath as Minerva, and AGI-Eval-MATH as AGI-M. The performance advantage of our method is preserved in both systems.}
    \label{tab:exp_other_controller}
    \vspace{1mm}
    \begin{tabular}{lcccccccc}
        \toprule
        \rowcolor{white}
        \textbf{Methods} & \textbf{M-LEval} & \textbf{MATH-500} & \textbf{Omni} & \textbf{Minerva} & \textbf{AGI-M} & \textbf{AMC23} & \makecell[c]{\textbf{AIME24}} & \textbf{Avg.} \\
        \cmidrule{2-9}
        \rowcolor{white}
        \multicolumn{9}{c}{\textbf{Controller: Qwen3-4B-Base}} \\
        \rowcolor{white}
        \multicolumn{9}{p{\dimexpr\linewidth-2\tabcolsep\relax}}{\textbf{Agent 1}: Qwen3-30B-A3B-Instruct; \textbf{Agent 2}: Qwen2.5-7B-Instruct; \textbf{Agent 3}: Qwen2.5-1.5B-Instruct.} \\
        \rowcolor{white}
        \midrule
        \rowcolor{white}
        Agent 1 & 0.956 & 0.938 & 0.551 & 0.828 & 0.933 & 0.975 & 0.833 & 0.859 \\
        Agent 2 & 0.757 & 0.750 & 0.254 & 0.524 & 0.734 & 0.500 & 0.100 & 0.517 \\
        Agent 3 & 0.554 & 0.542 & 0.157 & 0.273 & 0.547 & 0.275 & 0.067 & 0.345 \\
        \midrule
        Controller V1 & 0.761 & 0.738 & 0.255 & 0.460 & 0.736 & 0.550 & 0.133 & 0.519 \\
        Controller V2 & 0.883 & 0.804 & 0.328 & 0.419 & 0.809 & 0.775 & 0.333 & 0.622 \\
        Random Router & 0.756 & 0.744 & 0.304 & 0.537 & 0.737 & 0.575 & 0.275 & 0.561 \\
        RoBERTa Router  & 0.763 & 0.750 & 0.436 & 0.417 & 0.729 & 0.550 & 0.667 & 0.616 \\
        RouterDC & 0.784 & 0.764 & 0.445 & 0.419 & 0.736 & 0.600 & 0.275 & 0.575 \\
        Router-R1  & 0.811 & 0.808 & 0.501 & 0.480 & 0.798 & 0.900 & 0.767 & 0.724 \\
        \midrule
        \rowcolor{row-highlight}\textbf{Ours}  & \textbf{0.905}  & \textbf{0.878} & \textbf{0.535} & \textbf{0.665} & \textbf{0.878} & \textbf{0.925} & \textbf{0.833} & \textbf{0.803} \\
        \midrule
        \rowcolor{white}
        \multicolumn{9}{c}{\textbf{Controller: Qwen3-8B-Base}} \\
        \rowcolor{white}
        \multicolumn{9}{p{\dimexpr\linewidth-2\tabcolsep\relax}}{\textbf{Agent 1}: Qwen3-30B-A3B-Instruct; \textbf{Agent 2}: Ministral-3-8B-Instruct; \textbf{Agent 3}: Llama-3-2-1B-Instruct.} \\
        \midrule
        Agent 1 & 0.956 & 0.938 & 0.551 & 0.828 & 0.933 & 0.975 & 0.833 & 0.859 \\
        Agent 2 & 0.724 & 0.720 & 0.180 & 0.599 & 0.681 & 0.400 & 0.100 & 0.486 \\
        Agent 3 & 0.275 & 0.260 & 0.067 & 0.088 & 0.265 & 0.200 & 0.000 & 0.165 \\
        \midrule
        Controller V1 & 0.760 & 0.738 & 0.261 & 0.524 & 0.735 & 0.750 & 0.167 & 0.562 \\
        Controller V2 & 0.810 & 0.792 & 0.280 & 0.608 & 0.776 & 0.600 & 0.200 & 0.581 \\
        Random Router & 0.670 & 0.659 & 0.244 & 0.528 & 0.640 & 0.500 & 0.267 & 0.501 \\
        RoBERTa Router  & 0.659 & 0.662 & 0.458 & 0.376 & 0.628 & 0.775 & 0.833 & 0.627 \\
        RouterDC & 0.681 & 0.674 & 0.463 & 0.386 & 0.654 & 0.600 & 0.467 & 0.561 \\
        Router-R1  & 0.640 & 0.628 & 0.477 & 0.409 & 0.638 & 0.950 & 0.800 & 0.649 \\
        \midrule
        \rowcolor{row-highlight}\textbf{Ours} & \textbf{0.915} & \textbf{0.876} & \textbf{0.539} & \textbf{0.727} & \textbf{0.893} & \textbf{0.833} & \textbf{0.975} & \textbf{0.823} \\
       \bottomrule
    \end{tabular}
\end{table}

\subsection{Agent Usage Constraint and Sensitivity Analysis of the Lagrangian Multipliers}\label{sec:appen:lagrangian_sensitivity}

\textbf{Discussion on Agent Usage Constraints.} For routing in a multi-agent system with heterogeneous models, there exists an optimal partition of queries across agents. Ideally, queries that can be correctly handled by the weakest agent should be routed to that agent, and the remaining queries should be routed to stronger agents as needed. For a three-agent system, let the accuracies of the three agents be $p_1$, $p_2$, and $p_3$, where agent 3 is the weakest and agent 1 is the strongest. Then, the fraction of queries routed to agents 2 and 1 should be no larger than $1-p_3$ and $1-p_2$, respectively. This is the principle behind setting $\rho_1$ and $\rho_2$, which correspond to the usage constraints of agents 1 and 2. 
In our experiments, we treat these values as fixed system-level parameters chosen according to this principle.

\begin{figure}[t]
    \centering
    \begin{subfigure}[b]{0.48\textwidth}
        \centering
        \includegraphics[width=\textwidth]{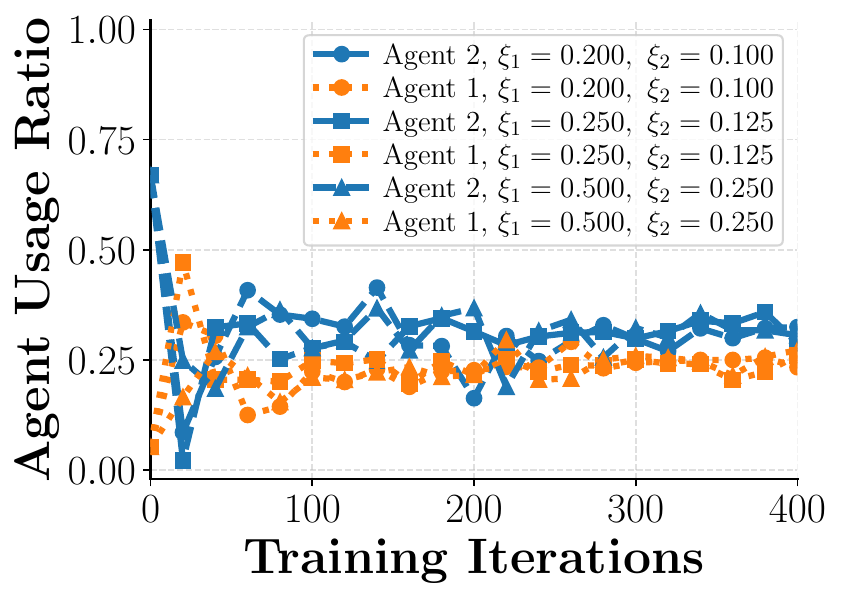}
        \caption{Agent usage under System 1.}
        \label{fig:ablate_lagrangian_system1}
    \end{subfigure}\hfill
    \begin{subfigure}[b]{0.48\textwidth}
        \centering
        \includegraphics[width=\textwidth]{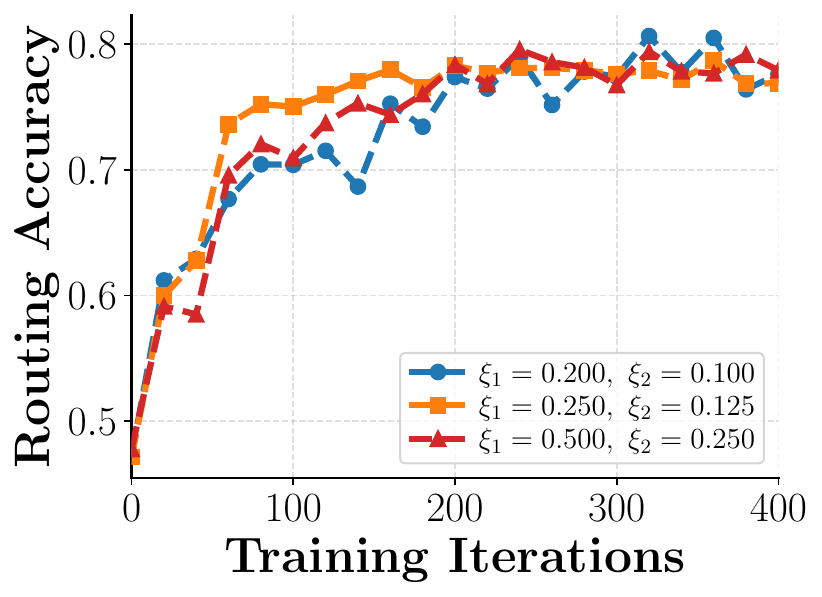}
        \caption{Routing accuracy under System 1.}
        \label{fig:ablate_lagrangian_routing_accuracy_system1}
    \end{subfigure}
    \caption{Sensitivity of our method to different Lagrangian multipliers.
    We vary the multiplier settings $(\xi_1, \xi_2)$ and report the resulting agent-usage and routing-accuracy dynamics. Across these choices, the learned controller remains stable and performs well: the usage ratios stay well controlled, while the routing accuracy remains consistently high after convergence.}
    \label{fig:ablate_lagrangian_sensitivity}
\end{figure}

\textbf{Sensitivity Analysis of the Lagrangian Multipliers.} We further evaluate the sensitivity of our method to the choice of Lagrangian multipliers. Specifically, we vary $(\xi_1,\xi_2)$ under System 1 and track both the agent-usage ratios and routing accuracy throughout training. Notably, in System 1, the controller is initialized with Qwen2.5-7B-Base, and the candidate agents are Qwen3-30B-A3B-Instruct, Qwen2.5-7B-Instruct, and
Qwen2.5-1.5B-Instruct.  
As shown in~\Cref{fig:ablate_lagrangian_sensitivity}, within the moderate range of multiplier values considered here, the training dynamics are broadly similar across settings: the usage ratios remain reasonably controlled, while the routing accuracy converges to a comparable level.
These results suggest that, within a moderate range of multiplier values, the controller can automatically learn a stable partition of queries across agents, while maintaining a favorable balance between satisfying the usage constraints and preserving routing performance.

\subsection{Limitation of One-Turn Routing}\label{sec:appen:one-turn}

To better understand the limitation of one-turn routing, we analyze its difficulty prediction behavior through a routing confusion matrix. In one-turn routing, the controller makes a one-shot estimate of the problem difficulty and then dispatches the query to one of three models accordingly. In System 1, Qwen2.5-1.5B-Instruct for easy problems, Qwen2.5-7B-Instruct for medium problems, and Qwen3-30B-A3B-Instruct for hard problems. In System 2, Llama-3-2-1B-Instruct for easy problems, Ministral-3-8B-Instruct for medium problems, and Qwen3-30B-A3B-Instruct for hard problems.  
The rows of the confusion matrix denote the ground-truth difficulty level, assigned according to the minimum-capability model that can correctly solve the problem: easy, medium, and hard correspond to problems solvable by the weakest, medium, and strongest model, respectively. The columns denote the model selected by the router. 
Correct routing appear on the diagonal, whereas off-diagonal entries indicate under-routing or over-routing errors. In particular, predicting a lower difficulty than the ground truth may route a problem to an insufficiently capable model, while predicting a higher difficulty than the ground truth may cause unnecessary reliance on stronger models.
The results are reported in~\Cref{tab:one_turn_routing_confusion_system1,tab:one_turn_routing_confusion_system2}. The controller is both trained and evaluated on the MATH-lighteval dataset, with a total of 5000 samples in the test split.

\begin{table}[h]
\centering
\small
\caption{Routing confusion matrix for one-turn routing in System 1. Rows denote the ground-truth problem difficulty, and columns denote the model selected by the router. The weakest, medium, and strongest models correspond to Qwen2.5-1.5B-Instruct, Qwen2.5-7B-Instruct, and Qwen3-30B-A3B-Instruct, respectively. Off-diagonal entries indicate routing errors that either under-estimate or over-estimate the required model capability.}
\setlength{\tabcolsep}{8pt}
\begin{tabular}{c|ccc}
\toprule
\multirow{2}{*}{\textbf{Problem Difficulty}} & \multicolumn{3}{c}{\textbf{Routed Model}} \\
\cmidrule(lr){2-4}
 & \textbf{Weakest} & \textbf{Medium} & \textbf{Strongest} \\
\midrule
\textbf{Easy}   & $2254$ & $302$ & $90$ \\
\textbf{Medium} & $\textbf{515}$ & $394$ & $228$ \\
\textbf{Hard}   & $\textbf{211}$ & $\textbf{330}$ & $676$ \\
\bottomrule
\end{tabular}
\label{tab:one_turn_routing_confusion_system1}
\end{table}
\begin{table}[h]
\centering
\small
\caption{Routing confusion matrix for one-turn routing in System 2. Rows denote the ground-truth problem difficulty, and columns denote the model selected by the router. The weakest, medium, and strongest models correspond to Llama-3-2-1B-Instruct, Ministral-3-8B-Instruct, and Qwen3-30B-A3B-Instruct, respectively. Off-diagonal entries indicate routing errors that either under-estimate or over-estimate the required model capability.}
\setlength{\tabcolsep}{8pt}
\begin{tabular}{c|ccc}
\toprule
\multirow{2}{*}{\textbf{Problem Difficulty}} & \multicolumn{3}{c}{\textbf{Routed Model}} \\
\cmidrule(lr){2-4}
 & \textbf{Weakest} & \textbf{Medium} & \textbf{Strongest} \\
\midrule
\textbf{Easy}   & $1064$ & $179$ & $49$ \\
\textbf{Medium} & $\textbf{1080}$ & $782$ & $468$ \\
\textbf{Hard}   & $\textbf{136}$ & $\textbf{248}$ & $994$ \\
\bottomrule
\end{tabular}
\label{tab:one_turn_routing_confusion_system2}
\end{table}

From~\Cref{tab:one_turn_routing_confusion_system1,tab:one_turn_routing_confusion_system2}, we observe that the router exhibits a systematic bias toward selecting weaker models. Notably, such bias, either toward weaker or stronger models, is inherent in one-shot routing, since perfectly partitioning queries by their required model capability is generally difficult. Under our resource constraints, where the usage of stronger models is explicitly limited, the router may further favor weaker models when the difficulty estimate is uncertain. This motivates our multi-turn coordination, which allows the controller to revise its routing decisions based on intermediate interactions.

\subsection{Controller's Capability in Verification}

Verification (i.e., evaluation) is a key component of our sequential coordination controller. In our multi-turn framework, re-routing occurs only when the controller identifies that the response produced in the current turn is not sufficiently reliable. Therefore, the controller must not only select an appropriate agent, but also judge whether the intermediate response should be accepted or further refined.

To evaluate this capability, we focus on the controller's verification behavior at the second turn, where the number of active queries is still sufficiently large before decreasing in later turns. We present confusion matrices over the two heterogeneous multi-agent systems considered in~\Cref{tab:exp_main}. In particular, here we do not verify responses generated by the strongest agent. In System 1 and System 2, the controllers are instantiated as Qwen2.5-7B-Base and Qwen3-4B-Base, respectively. Rows denote the ground-truth correctness of the current response, while columns denote the controller's verification decision. Correct verification decisions appear on the diagonal, whereas off-diagonal entries indicate false acceptance or false rejection. A false acceptance occurs when the controller incorrectly accepts an incorrect response, while a false rejection occurs when the controller rejects a correct response and continues the interaction unnecessarily. These results are reported in~\Cref{tab:confusion_verification_system1,tab:confusion_verification_system2}.

\begin{table}[h]
\centering
\caption{Confusion matrix of controller verification decisions in System 1. Controller is trained and tested on MATH-lighteval task.}
\label{tab:confusion_verification_system1}
\begin{tabular}{c c c}
\toprule
 & Predicted Incorrect & Predicted Correct \\
\midrule
Actual Incorrect & \textbf{892} & 271 \\
Actual Correct & 216 & 2805 \\
\bottomrule
\end{tabular}
\end{table}
\begin{table}[h]
\centering
\caption{Confusion matrix of controller verification decisions in System 1. Controller is trained and tested on MATH-lighteval task.}
\label{tab:confusion_verification_system2}
\begin{tabular}{c c c}
\toprule
 & Predicted Incorrect & Predicted Correct \\
\midrule
Actual Incorrect & \textbf{1247} & 298 \\
Actual Correct & 199 & 2563 \\
\bottomrule
\end{tabular}
\end{table}

From~\Cref{tab:confusion_verification_system1,tab:confusion_verification_system2}, we observe that the controller achieves high verification accuracy, with most entries concentrated on the diagonal. In particular, the number of false rejections remains relatively small, 216 in System 1 and 199 in System 2, indicating that the controller rarely discards correct responses. At the same time, the controller correctly identifies a large portion of incorrect responses, 892 in System 1 and 1247 in System 2, corresponding to incorrect-response recall rates of 76.69\% and 80.71\%. This demonstrates the controller's ability to filter out low-quality responses. Overall, the controller maintains meaningful rejection capability while preserving a high acceptance rate for correct responses. These results help explain why the multi-turn framework leads to improved performance: it enables the controller to identify insufficient intermediate responses and trigger re-routing.

\section{Prompt Template}\label{sec:appen:prompt}

\begin{table}[h]
\caption{\textbf{Controller prompt templates for multi-turn coordination.}
Template I corresponds to the initial-turn prompt $\mathbf{c}$, which performs routing based solely on the input query. Template II corresponds to the subsequent-turn prompt $\mathbf{c}^{\prime}$, which verifies the previous agent's solution and either terminates the interaction or routes the problem to a stronger agent.}
\label{tab:prompts}
\vspace{2mm}
\centering
\begin{tabular}{p{0.95\linewidth}}
\toprule
\textbf{Template I (initial routing prompt $\mathbf{c}$).}

\emph{System prompt:}
You are a router. 

Given a math problem, your task is to route the problem to the appropriate model based on its difficulty. 

Rules:
\begin{itemize}
\item The available models, ordered from weaker to stronger, are: \texttt{AGENT LIST}.
\item Choose the appropriate model based on the problem difficulty.
\end{itemize}

Output format:
\begin{itemize}
\item First, put your thinking process inside \texttt{<thinking>}...\texttt{</thinking>} tags.
\item Then, output your choice in the format \texttt{<model>}AGENT NAME\texttt{</model>}.
\item Your thinking should briefly explain the difficulty only.
\item You don't need to solve the problem.
\end{itemize}
\\
\midrule
\textbf{Template II (subsequent-turn verification and re-routing prompt $\mathbf{c}^{\prime}$).}

\emph{System prompt:}
You are a math solution verifier and router.

Given a math problem and a proposed solution from \texttt{AGENT NAME}, determine whether the proposed solution is fully correct.

Rules:
\begin{itemize}
\item Check the proposed solution step by step.
\item If any step is incorrect, unjustified, incomplete, or relies on invalid assumptions, return \texttt{False} and route the problem to a stronger model.
\item If the solution is fully correct, return \texttt{True} and stop.
\item The available models, ordered from weaker to stronger, are: \texttt{AGENT LIST}.
\item If routing is needed, select exactly one model that is stronger than \texttt{AGENT NAME}.
\end{itemize}

Output format (you MUST use exactly this structure):
\begin{itemize}
\item First, put your verification inside \texttt{<checking>}...\texttt{</checking>}.
\item Then output either \texttt{<verdict>}True\texttt{</verdict>} or \texttt{<verdict>}False\texttt{</verdict>}.
\item If and only if the verdict is \texttt{False}, output in the next line:
\texttt{<model>}AGENT NAME\texttt{</model>}
\end{itemize}
\\
\bottomrule
\end{tabular}
\end{table}

To instantiate the controller in~\Cref{sec:multi_turn_dynamics}, we design two system prompt templates provided in~\Cref{tab:prompts}. The initial-turn prompt $\mathbf{c}$ is used for routing at turn $t=1$, where the controller selects an agent based solely on the input query. The subsequent-turn prompt $\mathbf{c}^{\prime}$ is used for turns $t>1$, where the controller acts as both verifier and router: it evaluates the previous response and either terminates the interaction or re-routes the query to a stronger agent.

\section{Details on Datasets and Hyperparameters}\label{sec:appen:hyperparameter}

\textbf{Details on Datasets.} We fine-tune the controller on MATH-lighteval~\citep{hendrycks2measuring}. The MATH-lighteval dataset comprises 12500 problems drawn from mathematics competitions, covering topics such as algebra, geometry, counting and probability, number theory, and precalculus. We use 7500 samples (e.g., train split) for training and 5000 (e.g., test split) for testing. Additionally, we evaluate the system fine-tuned on the MATH-lighteval dataset against other 6 widely used mathematical benchmarks: MATH500~\citep{hendrycks2measuring},
Omni-MATH~\citep{gao2024omni}, MinervaMath~\citep{li2024numinamath},
AGI-Eval-MATH~\citep{zhong2024agieval}, AMC23~\citep{lewkowycz2022solving},
and AIME24.

\textbf{Hyperparameter Details.} The hyperparameters used by our method in the main experiments for Systems 1 and 2 are summarized in~\Cref{tab:hyperparameters}. These settings are used for the results reported in~\Cref{tab:exp_main} and~\Cref{fig:llm-training-dynamics}.

\begin{table}[H]
\caption{Hyperparameter settings used by our method in the main experiments for Systems 1 and 2, corresponding to the results in~\Cref{tab:exp_main} and~\Cref{fig:llm-training-dynamics}.}
\label{tab:hyperparameters}
\centering
\vspace{2mm}
\resizebox{1\linewidth}{!}{
\begin{tabular}{l|cc}
    \toprule
    \textbf{Hyperparameter} & System 1, controller: Qwen2.5-7B &  System 2, controller: Qwen3-4B\\
   \midrule
    Batch size $|\mathcal{D}_b|$ & 128 & 128 \\
   \midrule
    Group size $G$ & 4 & 4 \\
    \midrule
    Max prompt length & 3072 & 3072 \\ 
    \midrule
    Max response length & 2048 & 2048 \\
    \midrule
    Learning rate $\eta$ & 1e-6 & 1e-6 \\
    \midrule
    Total training iterations $S$ & 400 & 600 \\
    \midrule
    Sampling temperature for training & 1.0 & 1.0 \\
    \midrule
   Sampling temperature for evaluation & 0 & 0 \\ 
   \midrule
   Routing/Verification reward & 1 & 1 \\ 
   \midrule
   Discount factor $\gamma$ & 1 & 1 \\ 
   \midrule
   Maximum number of turns $T$ & 3 & 3 \\
   \midrule
   Agent usage constraint $\rho_1 / \rho_2 / \rho_3$ & 0.25/0.5/1 & 0.25/0.5/1 \\
   \midrule
   Lagrangian multiplier $\xi_1 / \xi_2 / \xi_3$ & 0.25/0.125/0 & 0.25/0.125/0 \\
    \bottomrule
\end{tabular}
}
\end{table}

\section{Proof of Proposition~\ref{Proposition_lagrangian_relaxation}}\label{sec:appen:proof_proposition_relaxation}

Noting that $\sum_{k=1}^K\sum_{t=1}^T \mathbb{E}[\mathbf{1}(f(\mathbf{a}_t)=k)] \leq T$. We first rewrite the constraint explicitly as:
\begin{equation}\label{equa:problem_formulation_relaxed_1}
    \begin{aligned}
        \underset{\boldsymbol{\theta}}{\text{maximize}} ~~ & \mathbb{E}_{\mathbf{q}\sim \mathcal{D}}\, \mathbb{E}_{\mathbf{a}_t \sim \pi_{\boldsymbol{\theta}}(\cdot \mid \mathbf{s}_t)}\!\left[\sum_{t=1}^{T} \gamma^{t-1} R(\mathbf{s}_t, \mathbf{a}_t)\right], \\
        \text{subject to} ~~ & \sum_{t=1}^T \mathbb{E}_{\mathbf{q}\sim \mathcal{D}}\, \mathbb{E}_{\mathbf{a}_t \sim \pi_{\boldsymbol{\theta}}(\cdot \mid \mathbf{s}_t)}\!\bigl[\mathbf{1}(f(\mathbf{a}_t)=k)\bigr] \leq \rho_k\, T, \quad \forall\, k.
    \end{aligned}
\end{equation}
Dualizing the constraints in \eqref{equa:problem_formulation_relaxed_1} yields the Lagrangian relaxation~\citep{schulman2015trust,schulman2017proximal}:
\begin{equation}\label{equa:problem_formulation_relaxed_2}
    \begin{aligned}
        \underset{\boldsymbol{\theta}}{\text{maximize}} ~~ & \mathbb{E}_{\mathbf{q}\sim \mathcal{D}}\, \mathbb{E}_{\mathbf{a}_t \sim \pi_{\boldsymbol{\theta}}(\cdot \mid \mathbf{s}_t)}\!\left[\sum_{t=1}^{T} \gamma^{t-1} R(\mathbf{s}_t, \mathbf{a}_t) \right] \\
        & - \sum_{k=1}^K \xi_k \!\left( \sum_{t=1}^T \mathbb{E}_{\mathbf{q}\sim \mathcal{D}}\, \mathbb{E}_{\mathbf{a}_t \sim \pi_{\boldsymbol{\theta}}(\cdot \mid \mathbf{s}_t)}\!\bigl[\mathbf{1}(f(\mathbf{a}_t)=k)\bigr] - \rho_k\, T \right),
    \end{aligned}
\end{equation}
where $\xi_k \geq 0$ is the Lagrange multiplier associated with the usage constraint on agent $k$, treated as a tunable hyperparameter. Dropping the constant term $\sum_{k=1}^K \xi_k \rho_k T$ and rearranging, we obtain:
\begin{equation}\label{equa:problem_formulation_relaxed_2b}
    \begin{aligned}
        \underset{\boldsymbol{\theta}}{\text{maximize}} ~~ & \mathbb{E}_{\mathbf{q}\sim \mathcal{D}}\, \mathbb{E}_{\mathbf{a}_t \sim \pi_{\boldsymbol{\theta}}(\cdot \mid \mathbf{s}_t)}\!\left[\sum_{t=1}^{T} \gamma^{t-1} R(\mathbf{s}_t, \mathbf{a}_t) - \sum_{t=1}^{T} \sum_{k=1}^K \xi_k\, \mathbf{1}(f(\mathbf{a}_t)=k)\right].
    \end{aligned}
\end{equation}
Under the reward definition in \eqref{reward_definition} and  $\gamma = 1$, Problem~\eqref{equa:problem_formulation_relaxed_2b} simplifies to:
\begin{equation}\label{equa:problem_formulation_relaxed_3}
\resizebox{0.93\linewidth}{!}{%
$ \displaystyle \underset{\boldsymbol{\theta}}{\text{maximize}} ~~ \mathbb{E}_{\mathbf{q}\sim \mathcal{D}}\, \mathbb{E}_{\mathbf{a}_t \sim \pi_{\boldsymbol{\theta}}(\cdot \mid \mathbf{s}_t)}\!\left[\sum_{t=1}^{T} \!\left(0.5 R_r(\mathbf{s}_t, \mathbf{a}_t) - \sum_{k=1}^K \xi_k\, \mathbf{1}(f(\mathbf{a}_t)\!=\!k) + 0.5 R_v(\mathbf{s}_t, \mathbf{a}_t) \right)\!\right] $.
}
\end{equation}
With the modified routing reward and composite reward defined in \eqref{modified_reward}, we arrive at the final unconstrained objective \eqref{equa:problem_formulation_relaxed_4}. This completes the proof of Proposition~\ref{Proposition_lagrangian_relaxation}.

\section{Algorithm Details}\label{sec:appen:algorithm}



\subsection{Problem~\eqref{equa:problem_formulation_relaxed_4} and GRPO}
\label{sec:appen:grpo}

\subsubsection{Policy Gradient of Problem~\eqref{equa:problem_formulation_relaxed_4}}

Problem~\eqref{equa:problem_formulation_relaxed_4} is a finite-horizon discounted MDP over controller states and actions, and the immediate reward is given by the modified reward $R'(\mathbf{s}_t,\mathbf{a}_t)$. Therefore, its objective is
\begin{equation}
J(\boldsymbol{\theta})
=
\mathbb{E}_{\mathbf{q}\sim\mathcal{D}}
\mathbb{E}_{\tau\sim \pi_{\boldsymbol{\theta}}}
\left[
\sum_{t=1}^{T}\gamma^{t-1}R'(\mathbf{s}_t,\mathbf{a}_t)
\right].
\end{equation}
It admits the standard policy-gradient form. In particular, fixing the query $\mathbf{q}$, we have~\citep{sutton1998reinforcement}
\begin{equation}
\nabla_{\boldsymbol{\theta}} J(\boldsymbol{\theta})
=
\mathbb{E}_{\tau\sim \pi_{\boldsymbol{\theta}}}
\left[
\sum_{t=1}^{T}
\nabla_{\boldsymbol{\theta}}
\log \pi_{\boldsymbol{\theta}}(\mathbf{a}_t\mid \mathbf{s}_t)\,
Q_t^{\pi}(\mathbf{s}_t,\mathbf{a}_t)
\right].
\label{equa:pg_q_form}
\end{equation}

Following the multi-turn interaction procedure in Section~\ref{sec:multi_turn_dynamics}, we introduce an additional index $b$ to distinguish different rollout prefixes under group sampling. For each prefix state $\mathbf{s}_t^{b}$, we sample a group of $G$ controller actions
$\{\mathbf{a}_t^{(b,i)}\}_{i=1}^{G}\sim \pi_{\boldsymbol{\theta}_{\mathrm{old}}}(\cdot\mid \mathbf{s}_t^{b})$.
Each sampled action yields an immediate modified reward
\begin{equation}
r_t^{(b,i)} = R'(\mathbf{s}_t^{b},\mathbf{a}_t^{(b,i)}),
\end{equation}
and induces a next state $\mathbf{s}_{t+1}^{(b,i)}$. We then recursively estimate the discounted return at state $\mathbf{s}_t^{b}$ as
\begin{equation}
\hat{V}_t(\mathbf{s}_t^{b})
=
\frac{1}{G}\sum_{i=1}^{G}
\left[
r_t^{(b,i)}+\gamma \hat{V}_{t+1}(\mathbf{s}_{t+1}^{(b,i)})
\right],
\qquad
\hat{V}_{T+1}(\cdot)=0.
\label{equa:recursive_group_value}
\end{equation}
Define the corresponding sample-based action-value estimate
\begin{equation}
\hat{Q}_t^{(b,i)}
=
r_t^{(b,i)}+\gamma \hat{V}_{t+1}(\mathbf{s}_{t+1}^{(b,i)}).
\label{equa:sample_q_estimate}
\end{equation}
Then a GRPO-style estimated advantage is obtained by subtracting the group average,
\begin{equation}
A_t^{(b,i)}
=
\hat{Q}_t^{(b,i)}
-
\frac{1}{G}\sum_{j=1}^{G}\hat{Q}_t^{(b,j)}.
\label{equa:group_relative_advantage_discounted}
\end{equation}

Using~\eqref{equa:group_relative_advantage_discounted}, we define the sample-based gradient estimator
\begin{equation}
\widehat{\nabla}_{\boldsymbol{\theta}} J(\boldsymbol{\theta})
=
\sum_{t=1}^{T}\frac{1}{G}\sum_{i=1}^{G}
\nabla_{\boldsymbol{\theta}}
\log \pi_{\boldsymbol{\theta}}(\mathbf{a}_t^{(b,i)}\mid \mathbf{s}_t^{b})
\,A_t^{(b,i)}.
\label{equa:pg_with_advantage}
\end{equation}
We next justify its relation to the standard policy gradient in~\eqref{equa:pg_q_form}. Substituting~\eqref{equa:group_relative_advantage_discounted} into~\eqref{equa:pg_with_advantage}, we obtain
\begin{equation}
\widehat{\nabla}_{\boldsymbol{\theta}} J(\boldsymbol{\theta})
=
\sum_{t=1}^{T}\frac{1}{G}\sum_{i=1}^{G}
\nabla_{\boldsymbol{\theta}}
\log \pi_{\boldsymbol{\theta}}(\mathbf{a}_t^{(b,i)}\mid \mathbf{s}_t^{b})
\left(
\hat{Q}_t^{(b,i)}
-
\frac{1}{G}\sum_{j=1}^{G}\hat{Q}_t^{(b,j)}
\right).
\end{equation}
Taking expectation conditioned on $\mathbf{s}_t^{b}$, and using that the $G$ actions are sampled i.i.d.\ from $\pi_{\boldsymbol{\theta}}(\cdot\mid \mathbf{s}_t^{b})$, the cross terms (i.e., $\nabla_{\boldsymbol{\theta}}
\log \pi_{\boldsymbol{\theta}}(\mathbf{a}_t^{(b,i)}\mid \mathbf{s}_t^{b})
\hat{Q}_t^{(b,j)}, \forall j\neq i$) vanish since $\nabla_{\boldsymbol{\theta}}
\log \pi_{\boldsymbol{\theta}}(\mathbf{a}_t^{(b,i)}\mid \mathbf{s}_t^{b})$ and $\hat{Q}_t^{(b,j)}$ are independent of each other and
\[
\mathbb{E}\!\left[\nabla_{\boldsymbol{\theta}}\log \pi_{\boldsymbol{\theta}}(\mathbf{a}_t^{(b,i)}\mid \mathbf{s}_t^{b})\mid \mathbf{s}_t^{b}\right]=0.
\]
Therefore, we have
\begin{equation}
\mathbb{E}\!\left[\widehat{\nabla}_{\boldsymbol{\theta}} J(\boldsymbol{\theta})\right]
=
\frac{G-1}{G}
\sum_{t=1}^{T}
\mathbb{E}\!\left[
\nabla_{\boldsymbol{\theta}}
\log \pi_{\boldsymbol{\theta}}(\mathbf{a}_t\mid \mathbf{s}_t)\,
\hat{Q}_t(\mathbf{s}_t,\mathbf{a}_t)
\right].
\label{equa:scaled_pg_estimator}
\end{equation}
Thus, the estimator in~\eqref{equa:pg_with_advantage} is proportional to the standard policy gradient, with proportionality constant $(G-1)/G$. Equivalently, if one wishes to match the standard policy gradient in expectation exactly, one may rescale the estimator by
$
\frac{G}{G-1},
$
yielding
\begin{equation}
\widehat{\nabla}_{\boldsymbol{\theta}}^{\,\mathrm{corr}} J(\boldsymbol{\theta})
=
\frac{G}{G-1}
\sum_{t=1}^{T}\frac{1}{G}\sum_{i=1}^{G}
\nabla_{\boldsymbol{\theta}}
\log \pi_{\boldsymbol{\theta}}(\mathbf{a}_t^{(b,i)}\mid \mathbf{s}_t^{b})
\,A_t^{(b,i)}.
\label{equa:pg_with_advantage_corrected}
\end{equation}
In practice, this constant only rescales the gradient magnitude and does not change the ascent direction, so it can be absorbed into the learning rate. For simplicity, we therefore use the uncorrected form in~\eqref{equa:pg_with_advantage}, which is fully aligned with the GRPO-style group-relative advantage used later.

\subsubsection{Policy Gradient in GRPO}

We next show that optimizing a GRPO surrogate with the advantage in~\eqref{equa:group_relative_advantage_discounted} is aligned with optimizing Problem~\eqref{equa:problem_formulation_relaxed_4}.

First, we consider the conventional single-turn form of GRPO. Following~\citep{liu2025prorl}, the simplified GRPO objective can be written as
\begin{equation}
    \begin{aligned}
        \max_{\boldsymbol{\theta}} ~& \mathbb{E}_{\mathbf{q}\sim \mathcal{D}, \{\mathbf{y}_i\}_{i=1}^G \sim \pi_{\boldsymbol{\theta}_{\text{old}}} ( \mathbf{q})}  \frac{1}{G} \sum_{i=1}^G \left\{\min \left[\frac{\pi_{\boldsymbol{\theta}}\left(\mathbf{y}_i \right)}{\pi_{\boldsymbol{\theta}_{\text{old}}}\left(\mathbf{y}_i \right)} A_{i},  \operatorname{clip}\left(\frac{\pi_{\boldsymbol{\theta}}\left(\mathbf{y}_i \right)}{\pi_{\boldsymbol{\theta}_{\text{old}}}\left(\mathbf{y}_i \right)}, 1-\varepsilon, 1+\varepsilon\right) A_{i} \right]  \right\}, \\
        ~&\text{where}~ A_{i} = (r_i - \text{mean}(\{r_i\}_{i=1}^G))/\text{std}(\{r_i\}_{i=1}^G).
    \end{aligned}
\end{equation}
By removing clipping (on-policy GRPO, $\boldsymbol{\theta}=\boldsymbol{\theta}_{\mathrm{old}}$) and standard deviation normalization, the GRPO objective reduces to
\begin{equation}
    \begin{aligned}
        \max_{\boldsymbol{\theta}}~ \mathbb{E}_{\mathbf{q}\sim \mathcal{D}, \{\mathbf{y}_i\}_{i=1}^G \sim \pi_{\boldsymbol{\theta}_{\text{old}}} ( \mathbf{q})}  \frac{1}{G} \sum_{i=1}^G \left[\frac{\pi_{\boldsymbol{\theta}}\left(\mathbf{y}_i \right)}{\pi_{\boldsymbol{\theta}_{\text{old}}}\left(\mathbf{y}_i \right)} A_{i} \right],~ \text{where}~ A_{i} = r_i - \text{mean}(\{r_i\}_{i=1}^G).
    \end{aligned}
\end{equation}

We cast it into multi-turn form under our notation:
\begin{equation}
\mathcal{L}^{\mathrm{GRPO}}(\boldsymbol{\theta})
=
\mathbb{E}
\left[\sum_{t=1}^T
\frac{1}{G}\sum_{i=1}^{G}
\frac{\pi_{\boldsymbol{\theta}}(\mathbf{a}_t^{(b,i)}\mid \mathbf{s}_t^{b})}
{\pi_{\boldsymbol{\theta}_{\mathrm{old}}}(\mathbf{a}_t^{(b,i)}\mid \mathbf{s}_t^{b})}
A_t^{(b,i)}
\right].
\end{equation}
At $\boldsymbol{\theta}=\boldsymbol{\theta}_{\mathrm{old}}$, we have the following unbiased gradient estimator for $\mathcal{L}^{\mathrm{GRPO}}(\boldsymbol{\theta})$:
\begin{equation}
\widehat{\nabla}_{\boldsymbol{\theta}}
\mathcal{L}^{\mathrm{GRPO}}(\boldsymbol{\theta})
=
\sum_{t=1}^{T}\frac{1}{G}\sum_{i=1}^{G}
\nabla_{\boldsymbol{\theta}}
\log \pi_{\boldsymbol{\theta}}(\mathbf{a}_t^{(b,i)}\mid \mathbf{s}_t^{b})
\,A_t^{(b,i)}.
\label{equa:grpo_gradient_form}
\end{equation}

As we can see, \eqref{equa:grpo_gradient_form} is exactly the same estimator as \eqref{equa:pg_with_advantage}. Therefore, optimizing the on-policy GRPO surrogate, without standard deviation normalization, yields the same update direction as optimizing our discounted multi-turn objective up to a constant factor $(G-1)/G$, which can be absorbed into the learning rate. Hence, our discounted multi-turn objective can be optimized using the standard GRPO framework after replacing the trajectory-level reward in conventional GRPO by the recursively computed discounted return estimate and removing the standard deviation normalization.


\subsection{Practical On-Policy GRPO Procedure for Problem~\eqref{equa:problem_formulation_relaxed_4}}\label{sec:appen:grpo_algorithm}

\begin{algorithm}[t]
\caption{
GRPO-Based Training for the Multi-Turn Controller
\label{alg:grpo_multi_turn}
}
\begin{algorithmic}[1]
\REQUIRE Controller policy $\pi_{\boldsymbol{\theta}}$, agent pool $\mathcal{M}$, training set $\mathcal{D}$, horizon $T$, group size $G$, discount factor $\gamma$, learning rate $\eta$
\FOR{iteration $=1,\ldots,S$}
    \STATE Sample a mini-batch $\mathcal{D}_b \subset \mathcal{D}$
    \FOR{each query $\mathbf{q} \in \mathcal{D}_b$}
        \STATE Initialize the rollout tree with root $\mathbf{s}_1^{(1)} \gets (\mathbf{q}, \mathbf{c})$ and active prefix set $\mathcal{B}_1 \gets \{()\}$
        \FOR{$t=1,\ldots,T$}
            \FOR{each prefix $b \in \mathcal{B}_t$}
                \STATE Sample $\{\mathbf{a}_t^{(b,i)}\}_{i=1}^{G} \sim \pi_{\boldsymbol{\theta}}(\cdot \mid \mathbf{s}_t^{b})$
                \FOR{$i=1,\ldots,G$}
                    \STATE Parse $m_t^{(b,i)} = f(\mathbf{a}_t^{(b,i)})$
                    \IF{$\mathbf{a}_t^{(b,i)}$ invokes an agent}
                        \STATE Query agent $m_t^{(b,i)}$ to obtain response $\mathbf{g}_t^{(b,i)}$
                        \STATE Set $\mathbf{s}_{t+1}^{(b,i)} \gets (\mathbf{q}, \mathbf{c}^{\prime}, \mathbf{g}_t^{(b,i)})$
                    \ELSE
                        \STATE Mark child $(b,i)$ as terminal
                    \ENDIF
                    \STATE Compute $r_t^{(b,i)} = R^{\prime}(\mathbf{a}_t^{(b,i)}, \mathbf{s}_t^{b})$
                \ENDFOR
            \ENDFOR
            \STATE Set $\mathcal{B}_{t+1} \gets \{(b,i): b\in\mathcal{B}_t,\ i=1,\ldots,G,\ \mathbf{s}_{t+1}^{(b,i)} \text{ is non-terminal}\}$
        \ENDFOR
        \STATE Set $\hat{V}_{T+1}(\cdot)\gets 0$
        \FOR{$t=T,\ldots,1$}
            \FOR{each prefix $b \in \mathcal{B}_t$}
                \STATE Compute $\hat{Q}_t^{(b,i)} = r_t^{(b,i)} + \gamma \hat{V}_{t+1}(\mathbf{s}_{t+1}^{(b,i)})$ for $i=1,\ldots,G$
                \STATE Compute $\hat{V}_t(\mathbf{s}_t^{b}) = \frac{1}{G}\sum_{i=1}^{G}\hat{Q}_t^{(b,i)}$
                \STATE Compute $A_t^{(b,i)} = \hat{Q}_t^{(b,i)} - \frac{1}{G}\sum_{j=1}^{G}\hat{Q}_t^{(b,j)}$ for $i=1,\ldots,G$
            \ENDFOR
        \ENDFOR
        \STATE Calculate the gradient $\widehat{\nabla} \mathcal{L}_{\mathbf{q}}^{\mathrm{GRPO}}(\boldsymbol{\theta})$ over the nodes $\{(\mathbf{s}_t^{b},\mathbf{a}_t^{(b,i)},A_t^{(b,i)})\}_{t,b,i}$ in the rollout traces $\mathcal{B}_T$, and scale it by $1/|\mathcal{B}_T|$, following~\eqref{equa:grpo_gradient_form} (or \eqref{equa:pg_with_advantage})
    \ENDFOR
    \STATE Update $\boldsymbol{\theta} \leftarrow \boldsymbol{\theta} + \eta \nabla_{\boldsymbol{\theta}}
    \frac{1}{|\mathcal{D}_b|}\sum_{\mathbf{q}\in\mathcal{D}_b} \widehat{\nabla}\mathcal{L}_{\mathbf{q}}^{\mathrm{GRPO}}(\boldsymbol{\theta})$
\ENDFOR
\end{algorithmic}
\end{algorithm}

The previous subsection establishes that, in our setting, the on-policy GRPO gradient is aligned with the policy gradient of Problem~\eqref{equa:problem_formulation_relaxed_4} when the conventional trajectory-level reward is replaced by the recursively computed discounted return and standard deviation normalization is omitted. We now summarize the resulting practical training procedure used in our method in~\Cref{alg:grpo_multi_turn}. For each query, training constructs a multi-turn rollout tree rooted at the initial state $\mathbf{s}_1=(\mathbf{q},\mathbf{c})$: each active prefix state $\mathbf{s}_t^{b}$ branches into $G$ sampled controller actions, the selected agent responses determine the child states, and the modified reward $R'(\mathbf{a}_t^{(b,i)}, \mathbf{s}_t^{b})$ is computed at every node. The algorithm then performs a backward recursion over the tree to compute $\hat{Q}_t^{(b,i)}$, $\hat{V}_t(\mathbf{s}_t^{b})$, and the group-relative advantages $A_t^{(b,i)}$, which are finally aggregated into the GRPO surrogate for updating $\boldsymbol{\theta}$.

\section{Case Study}
We present representative case studies illustrating how the controller coordinates multi-turn interactions with agents through verification, feedback, and re-routing decisions. These examples are based on System 1, where the controller is initialized from Qwen2.5-7B-Base and the candidate agents are Qwen3-30B-A3B-Instruct, Qwen2.5-7B-Instruct, and Qwen2.5-1.5B-Instruct.

{\scriptsize
\begin{tcolorbox}[colback=white!98!gray, breakable, enhanced, title=\large Case Study: Multi-Turn Controller-Agent Interaction]

    \begingroup
        \setlength{\fboxsep}{4pt}
        \colorbox{gray!15}{\makebox[\dimexpr\linewidth-2\fboxsep\relax][l]{\textbf{\large Example 1: Exit in 2 Turns}}}
    \endgroup
\vspace{10pt}

\textcolor{my_purple}{\textbf{Question:}} The points $(x, y)$ represented in this table lie on a straight line. The point $(13, q)$ lies on the same line. What is the value of $p + q?$ \\

\textcolor{dimred}{\textbf{Turn 1 (Controller):}} \texttt{<thinking>}This is a linear equation problem involving finding the slope and using it to determine the value of $p$ and \( q \). It is a medium-difficult problem because it requires understanding of linear equations and the concept of slope. The problem involves finding the slope between two points, using the slope to find the value of \( p \), and then using the value of \( p \) to find \( q \). Given the need for multiple steps and the application of algebraic concepts, I will route this problem to the Qwen2.5-7B-Instruct model, which is suitable for handling medium-difficult mathematical problems involving linear equations and algebraic manipulations.\texttt{</thinking>} 

\texttt{<model>}Qwen2.5-7B-Instruct\texttt{</model>} \\

\textcolor{my_green}{\textbf{Response from Qwen2.5-7B-Instruct:}} To determine the value of \( p + q \), we first need to find the equation of the line that passes through the given points. The points given are \((2, -5)\), \((p, -14)\), and \((p+2, -17)\).

First, we calculate the slope of the line using the points \((2, -5)\) and \((p, -14)\):
\[
\text{slope} = \frac{-14 - (-5)}{p - 2} = \frac{-9}{p - 2}
\]

Next, we use the points \((2, -5)\) and \((p+2, -17)\):
\[
\text{slope} = \frac{-17 - (-5)}{(p+2) - 2} = \frac{-12}{p}
\]

Since both expressions represent the slope of the same line,
\[
\frac{-9}{p - 2} = \frac{-12}{p}
\]

Cross-multiplying gives:
\[
-9p = -12(p - 2)
\]
\[
-9p = -12p + 24
\]
\[
3p = 24
\]
\[
p = 8
\]

Using the point \((2, -5)\) and slope \(-\frac{3}{2}\), the equation is:
\[
y - (-5) = -\frac{3}{2}(x - 2)
\]
\[
y + 5 = -\frac{3}{2}x + 3
\]
\[
y = -\frac{3}{2}x - 2
\]

When \(x = 13\),
\[
q = -\frac{3}{2}(13) - 2 = -\frac{39}{2} - \frac{4}{2} = -\frac{43}{2}
\]

Finally,
\[
p + q = 8 - \frac{43}{2} = \frac{16}{2} - \frac{43}{2} = -\frac{27}{2}
\]

Thus, the value of \(p + q\) is \(\boxed{-\frac{27}{2}}\). \\

\textcolor{dimred}{\textbf{Turn 2 (Controller):}} \texttt{<checking>} Let's trace the proposed solution step by step and verify its correctness.

1. Calculate the slope using the points (2, -5) and (p, -14): 

\[\text{slope} = \frac{-14 - (-5)}{p - 2} = \frac{-9}{p - 2}\]

2. Calculate the slope using the points (2, -5) and (p+2, -17):

\[\text{slope} = \frac{-17 - (-5)}{(p+2) - 2} = \frac{-12}{p} \] 

3. Set the slopes equal to each other and solve for \( p \):

\[\frac{-9}{p - 2} = \frac{-12}{p}\]

Cross-multiplying gives:

\[-9p = -12(p - 2) \]    

\[-9p = -12p + 24 \]  

\[   3p = 24 \]   

\[p = 8 \] 

4. Verify the equation of the line using the point (2, -5) and the slope \(-\frac{3}{2}\):

\[ y + 5 = -\frac{3}{2}(x - 2) \] 

\[ y = -\frac{3}{2}x + 3 - 5 \]

\[ y = -\frac{3}{2}x - 2 \] 

5. Find the value of \( q \) when \( x = 13 \):

\[ q = -\frac{3}{2}(13) - 2 = -\frac{39}{2} - 2 = -\frac{39}{2} - \frac{4}{2} = -\frac{43}{2} = -21.5 \] 

6. Calculate \( p + q \):  \[ p + q = 8 + (-21.5) = -13.5 \] 

\begin{verbatim}
</checking>

<verdict>True</verdict>
\end{verbatim}

\vspace{15pt} 
    \begingroup
        \setlength{\fboxsep}{4pt}
        \colorbox{gray!15}{\makebox[\dimexpr\linewidth-2\fboxsep\relax][l]{\textbf{\large Example 2: Exit in 3 Turns}}}
    \endgroup
\vspace{10pt}

\textcolor{my_purple}{\textbf{Question:}} Rationalize the denominator: $\frac1{2\sqrt7}$. \\

\textcolor{dimred}{\textbf{Turn 1 (Controller):}} \texttt{<thinking>}This is a simple problem involving rationalizing the denominator. It is a basic algebraic manipulation problem that does not require advanced mathematical knowledge or complex problem-solving skills. Therefore, I will route this problem to the Qwen2.5-1.5B-Instruct model, which is suitable for easy problems.\texttt{</thinking>} 

\texttt{<model>}Qwen2.5-1.5B-Instruct\texttt{</model>} \\

\textcolor{my_green}{\textbf{Response from Qwen2.5-1.5B-Instruct:}} To rationalize the denominator of the expression $\frac{1}{2\sqrt{7}}$, we need to eliminate the square root from the denominator. We can do this by multiplying both the numerator and the denominator by the conjugate of the denominator. The conjugate of $2\sqrt{7}$ is $2\sqrt{7}$ itself, because it is the same number. 

Step 1: Multiply the numerator and the denominator by $2\sqrt{7}$

$$\frac{1}{2\sqrt{7}} \cdot \frac{2\sqrt{7}}{2\sqrt{7}}$$

Step 2: Simplify the expression. 
$$\frac{2\sqrt{7}}{2\sqrt{7} \cdot 2\sqrt{7}}$$ 

Step 3: Simplify the denominator. 

$$\frac{2\sqrt{7}}{2 \cdot 7}$$ 

Step 4: Simplify the fraction. 

$$\frac{2\sqrt{7}}{14}$$ 

Step 5: Simplify further by dividing both the numerator and the denominator by 2.

$$\frac{\sqrt{7}}{7}$$ 

Therefore, the rationalized form of the expression is $\boxed{\frac{\sqrt{7}}{7}}$. \\

\textcolor{dimred}{\textbf{Turn 2 (Controller):}} \texttt{<checking>} Let's trace the proposed solution step by step and verify its correctness.  

1. The given expression is \(\frac{1}{2\sqrt{7}}\). 

2. To rationalize the denominator, we multiply the numerator and the denominator by \(2\sqrt{7}\). 

3. This gives us \(\frac{2 \sqrt{7}}{2\sqrt{7} \cdot 2\sqrt{7}} = \frac{\sqrt{7}}{2 \cdot 7} = \frac{\sqrt{7}}{14}\).

4. The proposed solution seems to have an error in step 4.

We need to route the problem to a stronger agent to refine the solution. Qwen2.5-7B-Instruct is good choice, which is stronger than Qwen2.5-1.5B-Instruct.

\begin{verbatim}
</checking>

<verdict>False</verdict>

<model>Qwen2.5-7B-Instruct</model>
\end{verbatim}

\textcolor{my_green}{\textbf{Response from Qwen2.5-7B-Instruct:}} To rationalize the denominator of the expression $\frac{1}{2\sqrt{7}}$, we eliminate the square root from the denominator as follows:

\textbf{Identify the denominator:} The denominator is $2\sqrt{7}$.

\textbf{Multiply by a suitable factor:} To rationalize the denominator, multiply both the numerator and denominator by $\sqrt{7}$:
    \[
    \frac{1}{2\sqrt{7}} \cdot \frac{\sqrt{7}}{\sqrt{7}}
    \]
    
\textbf{Simplify the expression:}
    \[
    \frac{1 \cdot \sqrt{7}}{2\sqrt{7} \cdot \sqrt{7}} = \frac{\sqrt{7}}{2 \cdot 7}
    \]

\textbf{Simplify the denominator:}
    \[
    2 \cdot 7 = 14
    \]
    
\textbf{Final result:}
    \[
    \frac{\sqrt{7}}{14}
    \]

Thus, the rationalized form of $\frac{1}{2\sqrt{7}}$ is 
\[
\boxed{\frac{\sqrt{7}}{14}}.
\] \\

\textcolor{dimred}{\textbf{Turn 3 (Controller):}} \texttt{<checking>} Let's trace the proposed solution step by step and verify its correctness.  

1. The given expression is \(\frac{1}{2\sqrt{7}}\). 

2. To rationalize the denominator, we multiply the numerator and the denominator by \(\sqrt{7}\). 

3. This gives us \(\frac{\sqrt{7}}{2\sqrt{7} \cdot \sqrt{7}} = \frac{\sqrt{7}}{2 \cdot 7} = \frac{\sqrt{7}}{14}\).

4. The proposed solution is correct.

\begin{verbatim}
</checking>

<verdict>True</verdict>
\end{verbatim}

\end{tcolorbox}
}

\end{document}